\documentclass[12pt, onecolumn]{article}

\usepackage[margin=1.0in]{geometry}
\usepackage[hidelinks, hypertexnames=false]{hyperref}   
\usepackage{enumitem}
\usepackage[table]{xcolor}
\usepackage{mathtools, amsthm}
\usepackage[utf8]{inputenc}
\usepackage{float}
\usepackage{subcaption}
\usepackage{listings}
\usepackage{amsmath}
\usepackage{graphicx}
\usepackage{newtxmath}
\usepackage{url}
\usepackage[hidelinks, hypertexnames=false]{hyperref}
\usepackage[numbers]{natbib}
\usepackage{caption}
\usepackage{mathtools}
\usepackage{subcaption}
\usepackage{bbm}
\usepackage{acro}
\usepackage{booktabs} 
\usepackage{relsize}
\usepackage{multicol}
\usepackage{multirow}
\usepackage{booktabs} 
\usepackage{makecell}

\usepackage{lipsum}

\newtheorem{theorem}{Theorem}[section]

\newtheorem{lemma}[theorem]{Lemma}

\theoremstyle{definition}

\newcommand\restr[2]{{
  \left.\kern-\nulldelimiterspace 
  #1 
  \vphantom{\big|} 
  \right|_{#2} 
  }}

\newcommand{\norm}[1]{\left\lVert#1\right\rVert}

\makeatletter
\def\munderbar#1{\underline{\sbox\tw@{$#1$}\dp\tw@\z@\box\tw@}}
\makeatother

\AtBeginDocument{
  }

\renewenvironment{abstract}{
  \begin{center}
    \bfseries \abstractname\vspace{-.5em}
  \end{center}
  \normalsize
}{}

\theoremstyle{theorem}
\newtheorem{definition}{Definition}[section]

\allowdisplaybreaks

\title{\textbf{Signature Maximum Mean Discrepancy Two-Sample Statistical Tests}}

\author{\vspace{-0.35em} \textbf{Andrew Alden} \\\vspace{-0.35em}
  \small Department of Informatics\\\vspace{-0.35em}
  \small King's College London\\\vspace{-0.35em}
  \small \texttt{andrew.alden@kcl.ac.uk}
  \and
  \vspace{-0.35em} \textbf{Blanka Horvath} \\\vspace{-0.35em}
  \small Department of Mathematics \\\vspace{-0.35em}
  \small University of Oxford \\\vspace{-0.35em}
  \small \texttt{blanka.horvath@maths.ox.ac.uk}
  \and
  \vspace{-0.35em} \textbf{Zacharia Issa} \\\vspace{-0.35em}
  \small Department of Mathematics\\\vspace{-0.35em}
  \small King's College London \\\vspace{-0.35em}
  \small \texttt{zacharia.issa@kcl.ac.uk} \\\vspace{-0.35em}
}
\date{}

\usepackage[width=\textwidth]{caption}

\begin{document}

\maketitle

\abstract{Maximum Mean Discrepancy (MMD) is a widely used concept in machine learning research which has gained popularity in recent years as a highly effective tool for comparing (finite-dimensional) distributions. Since it is designed as a kernel-based method, the MMD can be extended to path space valued distributions using the signature kernel. The resulting signature MMD (sig-MMD) can be used to define a metric between distributions on path space. Similarly to the original use case of the MMD as a test statistic within a two-sample testing framework, the sig-MMD can be applied to determine if two sets of paths are drawn from the same stochastic process. This work is dedicated to understanding the possibilities and challenges associated with applying the sig-MMD as a statistical tool in practice. We introduce and explain the sig-MMD, and provide easily accessible and verifiable examples for its practical use. We present examples that can lead to Type 2 errors in the hypothesis test, falsely indicating that samples have been drawn from the same underlying process (which generally occurs in a limited data setting). We then present techniques to mitigate the occurrence of this type of error.}

\section{Introduction}
\label{sec:intro}
The seminal paper \textit{A Kernel Two-Sample Test} \cite{first_order_mmd_distinguish} lays the foundation for establishing a versatile and robust framework for analysing and comparing distributions. This framework has been assessed and benchmarked against established classical two-sample tests in the finite-dimensional setting (such as Kolmogorov-Smirnov test) where it demonstrated superior or comparable performance \cite{sig_mom_charac_law, mmd_null, first_order_mmd_distinguish}. It has also been shown to be well-suited for machine learning (ML) applications \cite{conference_paper, mmd_bayes, issa2023non, dist_reg_paper, rl_mmd}. This framework is therefore particularly well-known for its applications to analysing finite-dimensional distributions in a ML context. \newline

The test statistic which underpins this framework is based on computing the largest average difference between distributions, over functions in the unit ball of a Reproducing Kernel Hilbert Space (RKHS). This concept has been named the kernel Maximum Mean Discrepancy (MMD) \cite{first_order_mmd_distinguish}. On an appropriate (rich enough) space, the kernel MMD determines a metric between probability distributions. Using a signature kernel ($k_{\text{Sig}}$) (cf \cite{general_sig_kernels, sig_mom_charac_law, kernel_sequential_data, sig_kernel_pde_paper}), the aforementioned MMD can be established for measures on path space defined over the set $\mathcal{X}_{1-\text{var}}$. Similar to the originally proposed framework of the MMD (cf \cite{first_order_mmd_distinguish}), its path-valued counterpart (sig-MMD) can also be applied as a test statistic in a two-sample hypothesis test to determine if two collections of time series are drawn from different stochastic processes. If $\mathbf{X}$ and $\mathbf{Y}$ are two stochastic processes with laws $\mathbb{P}_{\mathbf{X}} = \mathbb{P} \circ \mathbf{X}^{-1}$ and $\mathbb{P}_{\mathbf{Y}} = \mathbb{P} \circ \mathbf{Y}^{-1}$ respectively, the two-sample test is described by the following null hypothesis ($H_{0}$) and alternative hypothesis ($H_{1}$):
\begin{equation*}
    H_{0} \colon \mathbb{P}_{\mathbf{X}} = \mathbb{P}_{\mathbf{Y}}~\text{against}~H_{1} \colon \mathbb{P}_{\mathbf{X}} \neq \mathbb{P}_{\mathbf{Y}}.
\end{equation*}
The sig-MMD has facilitated the use of distribution regression on path space \cite{conference_paper, american_pricing, higher_order_kme_neurips} and the training of generative models for time series \cite{vae_paper, issa2023non}. \newline

Since the path signature is infinite dimensional, the extension of these concepts to path space using the (full) signature within numerical algorithms is non-trivial. Recent literature focuses on the numerical computation of the signature kernel. These include (i) truncating the signature \cite{kernel_sequential_data}, (ii) using a kernel trick by means of an associated partial differential equation (PDE) to work with the full signature \cite{sig_kernel_pde_paper, higher_order_kme_neurips}, and (iii) using random Fourier features \cite{rff_sig, rffs}. This work aims to demonstrate how to use the sig-MMD in practice with specific focus on effectively navigating the challenges that may emerge in this context. We give practical insights for setting up the associated ML algorithms that can save the interested user considerable time when implementing these tools. We also suggest hyperparameter settings and discuss common pitfalls. We provide simple and easy-to-verify examples together with explanatory discussions which help build intuition surrounding these tools. \newline

As mentioned in \cite{sig_ker_chap}, in practice, the numerical approximation of the signature kernel is based on various hyperparameters. We study the statistical power of the two-sample hypothesis test as a function of the following subset of hyperparameters:
\begin{itemize}
    \item \textbf{Static Kernel Parameters} $\Theta_{k}$: Instead of working directly with the original paths, the paths can be embedded in an infinite dimensional feature space using another kernel \cite{kernel_sequential_data, sig_kernel_pde_paper}.
    \item \textbf{Truncation Parameter} $\mathbb{N}_{\infty}$: Since the signature of a path is an infinite dimensional object, it is not possible to work with the full signature. To work directly with the signature, one must truncate the signature by using the first $M \in \mathbb{N}$ terms of the signature. However, when using kernel methods, a kernel trick can be used to work with the full signature. In this case, computing the signature kernel amounts to solving a PDE \cite{higher_order_kme_neurips}. Therefore, when using the signature, one works with either an imperfect representation of the signature through its truncated version or the full signature by way of the kernel trick.
    \item \textbf{Preprocessing Parameters} $\Theta_{\text{preprocess}}$: Preprocessing the paths by using lead-lag transformations \cite{sig_ml, sig_ml_2, hoff_lead_lag, sig_ml_3}, adding a time-component \cite{sig_ml, sig_ml_3}, and by standardising dimensions\footnote{If the path is multi-dimensional, this involves standardising the individual dimensions. This could be done by subtracting the mean to have each dimension centred around 0.} \cite{issa2023non} can be performed.
    \item \textbf{Normalisation Parameters} $\Theta_{\text{normalisation}}$: Normalising the signature kernel results in a more robust statistic \cite{sig_mom_charac_law}. Moreover, by rescaling inner products, a generalised form of the signature kernel is obtained through which further signature MMDs can be constructed \cite{general_sig_kernels, weighted_sig}. 
    \item \textbf{Time Discretisation Parameters} $\Theta_{\text{discrete}}$: In practice, computations are performed on discrete data. The discretisation of the time interval can sometimes be controlled. Finer grids provide better approximations of the stochastic process \cite{donsker} whilst also increasing the computational burden. In addition, if working with the full signature through the PDE approach, the PDE is solved on a discrete grid which itself is controlled through a hyperparameter. 
\end{itemize}

The following sections are aimed at quantifying and interpreting the behaviour of the sig-MMD test statistic in the context of the two-sample hypothesis test. This behaviour is mostly characterised by the above parameters and therefore a discussion of the role of these parameters in improving performance on the hypothesis test is provided. The variance of the sig-MMD plays a significant role in characterising this behaviour. This variance is closely linked with the number of sample paths used. Although this can sometimes be directly addressed through sampling more samples from the processes and/or collecting additional data if the paths correspond to a physical system, one cannot disregard the computational bottlenecks associated with the sig-MMD. Therefore, increasing the sample size is generally not a feasible solution. We provide insights into stabilising the errors associated with the sig-MMD in the two-sample hypothesis testing framework in a high variance setting. \newline

Specifically, we focus on the following:
\begin{enumerate}
    \item \textbf{Hypothesis Testing Errors Associated with the Sig-MMD}: In most cases, the standard two-sample test using the sig-MMD on the space of stochastic processes has a high probability of a Type 2 error (acceptance of $H_0$ when $H_1$ is true) \cite{sig_mmd_example}. We explore techniques for reducing this. We show that scaling of the input paths plays a critical role in stabilising the statistical test. The effect of scaling on the signature kernel was first studied in \cite{general_sig_kernels, weighted_sig}. They showed that this has the equivalent effect of weighting inner products\footnote{The choice of path scaling is contained within the set $\Theta_{\text{normalisation}}$.} \cite[Lemma 2.9]{general_sig_kernels}, \cite[Lemma 2.10]{weighted_sig}. By weighting inner products, we can zoom in and focus on specific moments of the processes and tailor the two-sample hypothesis test to the specific processes. We leverage the relationship between path scaling and the decaying absolute value of the graded signature terms to construct a two-sample test statistic targeted at improving the test power. In practice, the true distributions of the stochastic processes are not observed and instead, empirical distributions are used to approximate the true distributions. Since sample sizes affect the quality of the approximation, we study the likelihood of Type 2 errors occurring across various sample sizes. We also study the effect of path scaling on the likelihood of a Type 1 error occurring. 
    \item \textbf{Information Associated with the Signature Level Terms}: We study the effects the different levels of the signature have on the statistical power of the two-sample test. In particular, we focus on the trade-off between the level of truncation ($\mathbb{N}_{\infty}$) and the information embedded within the higher-order terms. Using the kernel trick allows us to study the effects of including the full signature when computing the MMD. We then need to account for the factorial decay of the signature terms to ensure that higher level terms are contributing to the final value of the MMD. 
\end{enumerate}

\noindent
\textbf{Acknowledgements} This work was supported by the UK Engineering and Physical Sciences Research Council (EPSRC) under Grant [EP/T517963/1]. The code is publicly available and can be found at the repository: \url{https://github.com/Andrew-Alden/SignatureMMDTesting}.

\section{General Signature MMD}\label{sec:fact_decay}

The path signature is well suited for working with data streams as a feature map for learning tasks. As a transform, the signature is invariant to reparametrisation, filtering out symmetries \cite{sig_kernel_pde_paper}, and a continuous real-valued function on path space can be approximated by a linear function of the signature arbitrarily well \cite{sig_ml_2, kernel_sequential_data}. This result is analogous to the classical result that a continuous real-valued function defined on a closed interval can be approximated arbitrarily well by a polynomial function. In our case, the signature corresponds to a polynomial on path space. In terms of learning tasks, instead of fitting a non-linear function on path space, a linear function (i.e.\ linear regression) with the (truncated) signature terms as regressors can be used. \newline

When a path $\mathbf{x}$ has finite $p$-variation ($p \geq 1$), the signature terms ($\Phi_{\text{Sig}, m} \left(\mathbf{x}\right)$) decay factorially according to the uniform estimate \cite{sig_kernel_pde_paper}, \cite[Theorem 3.2]{sig_ml_3}, \cite[Lemma 5.1]{sig_defn_paper_2} 
\begin{equation}\label{eqn:sig_factorial_decay}
    \left| \left| \Phi_{\text{Sig}, m} \left(\mathbf{x}\right) \right| \right| \leq \frac{\left| \left| \mathbf{x} \right| \right|_{p}^{m}}{m!}
\end{equation}
where $\left| \left| \cdot \right| \right|_{p}$ denotes the $p$-variation metric. Computationally, working with the signature is challenging as it is infinite dimensional, and so, a large proportion of signature-based ML approaches use the truncated signature. In terms of absolute value, only a small amount of information is lost by truncating due to the factorial decay \cite{dist_reg_paper, sig_wgan, sig_ml_3}. However, we show that higher-order terms play a critical role in capturing distribution differences through the sig-MMD. In a regression setting, specific signature terms can be re-weighted to account for the factorial decay. In the case of the sig-MMD, this is a non-trivial task. This plays a key role in this work. 

\subsection{Sig-MMD}

Let $H$ be a RKHS associated with kernel $k$. Also, let $\mathcal{F}$ be a unit ball in $H$ defined over a compact metric space $\mathcal{X}$. Given two distributions $\mu, \nu \in \mathcal{P} \left(\mathcal{X}\right)$, the MMD between $\mu$ and $\nu$ is defined as \cite{first_order_mmd_distinguish}
\begin{equation*}
    d_{k} \left(\mu, \nu \right) \coloneqq \sup_{f \in \mathcal{F}} \left| \int_{\mathcal{X}} f \left(\mathbf{x}\right) \mu \left(d \mathbf{x}\right) - \int_{\mathcal{X}} f \left( \mathbf{x}\right) \nu \left(d \mathbf{x}\right) \right|. 
\end{equation*}
If $k$ is a continuous and characteristic kernel, then $d_{k}$ is a metric \cite[Theorem 5]{first_order_mmd_distinguish}. If $H$ is the RKHS associated with the signature kernel $k_{\text{Sig}}$\footnote{Given $\mathbf{x}, \mathbf{y} \in \mathcal{X}_{1-\text{var}}$ the signature kernel is defined as $k_{\text{Sig}} \left(\mathbf{x}, \mathbf{y}\right) \coloneqq \langle \Phi_{\text{Sig}} \left(\mathbf{x}\right), \Phi_{\text{Sig}} \left(\mathbf{y}\right) \rangle$.}, the corresponding MMD is the sig-MMD. Since the signature kernel is characteristic \cite{sig_mom_charac_law, higher_order_kme_neurips} and universal \cite{sig_mom_charac_law} when restricted to a compact subset $\mathcal{K} \subset \mathcal{X}_{1-\text{var}}$, the sig-MMD is a metric on the space of distributions of stochastic processes defined over paths in $\mathcal{K}$. Throughout this work, $\mathcal{K} \subset \mathcal{X}_{1-\text{var}}$ denotes a compact subset of $\mathcal{X}_{1-\text{var}}$.\newline

Throughout this work, we consider processes in continuous time. However, in practice, we only have access to discrete datapoints. A path is constructed from the data using linear interpolation. The linear interpolated paths are of bounded variation and belong to the set $\mathcal{X}_{1-\text{var}}$. For this reason, throughout this work we assume that all paths are elements of $\mathcal{X}_{1-\text{var}}$ even though their continuous counterpart may not be. \newline

Given two stochastic processes $\mathbf{X}$ and $\mathbf{Y}$, to compute the sig-MMD between $\mathbb{P}_{\mathbf{X}}$ and $\mathbb{P}_{\mathbf{Y}}$, it is first re-formulated in terms of Kernel Mean Embeddings (KMEs). The KME of a stochastic process maps the distribution of the stochastic process to an element in a RKHS. Let $H_{\text{Sig}}$ be the RKHS with kernel $k_{\text{Sig}}$ restricted to paths in the compact subset $\mathcal{K}$. Since the signature kernel is characteristic when restricted to the compact subset $\mathcal{K}$, each distribution defined over paths in $\mathcal{K}$ is embedded as a unique element within the RKHS $H_{\text{Sig}}$. Formally, the KME\footnote{Further details on KMEs can be found in \cite{first_order_mmd_distinguish, kme_review, higher_order_kme_neurips, kme_article}.} is given by the mapping\footnote{Since for practical purposes we work with elements from the set $\mathcal{X}_{\text{Seq}}$, the mapping $\mu$ extends to this set by means of the natural inclusion $\mathcal{X}_{\text{Seq}} \hookrightarrow \mathcal{K} \subset \mathcal{X}_{1-\text{var}}$ induced through linear interpolation.}
\begin{equation*}
    \mu_{\mathbf{X}} \colon \mathcal{P} \left( \mathcal{K} \right) \mapsto H_{\text{Sig}}
\end{equation*}
defined by
\begin{equation*}
    \mu_{\mathbf{X}} \colon \mathbb{P}_{\mathbf{X}} \mapsto \mathbb{E}_{\mathbf{X} \sim \mathbb{P}_{\mathbf{X}}} \left[k_{\text{Sig}} \left(\mathbf{X}, \cdot \right) \right]. 
\end{equation*}
Then, the sig-MMD is given by \cite[Lemma 4]{first_order_mmd_distinguish}
\begin{align} \label{eqn:mmd_norm}
    \nonumber d_{k_{\text{Sig}}} \left(\mathbb{P}_{\mathbf{X}},  \mathbb{P}_{\mathbf{Y}} \right)^{2} &=\norm{\mu_{\mathbf{X}} - \mu_{\mathbf{Y}}}_{H_{\text{Sig}}}^{2}  \\
    &= \norm{\mathbb{E}_{\mathbf{X} \sim \mathbb{P}_{\mathbf{X}}} \left[ k_{\text{Sig}} \left(\mathbf{X}, \cdot\right) \right] - \mathbb{E}_{\mathbf{Y} \sim \mathbb{P}_{\mathbf{Y}}} \left[ k_{\text{Sig}} \left(\cdot, \mathbf{Y}\right) \right]}_{H_{\text{Sig}}}^{2}
\end{align}
\noindent
with $\norm{\cdot}_{H_{\text{Sig}}}$ being the norm induced by the inner product of the RKHS $H_{\text{Sig}}$. \newline

The KME formulation provided in Eq.\ \eqref{eqn:mmd_norm} shows that the sig-MMD is the norm of the difference between the embeddings of the laws of the stochastic processes. Given independent collections $\mathbf{X}, \mathbf{X}^{\prime} \sim \mathbb{P}_{\mathbf{X}}$ and $\mathbf{Y}, \mathbf{Y}^{\prime} \sim \mathbb{P}_{\mathbf{Y}}$, by expanding the norm into inner products and using the reproducing property of the kernel operator, an alternative formulation of the sig-MMD is given by \cite[Lemma 6]{first_order_mmd_distinguish}
\begin{eqnarray} \label{eqn:mmd_kernel}
    \nonumber d_{k_{\text{Sig}}} \left(\mathbb{P}_{\mathbf{X}}, \mathbb{P}_{\mathbf{Y}} \right)^{2} &=& \mathbb{E}_{\mathbf{X}, \mathbf{X}^{\prime} \sim \mathbb{P}_{\mathbf{X}}} \left[ k_{\text{Sig}} \left(\mathbf{X}, \mathbf{X}^{\prime}\right) \right]  + \mathbb{E}_{\mathbf{Y}, \mathbf{Y}^{\prime} \sim \mathbb{P}_{\mathbf{Y}}} \left[k_{\text{Sig}} \left(\mathbf{Y}, \mathbf{Y}^{\prime}\right) \right]  \\
    &-& 2 \mathbb{E}_{\mathbf{X} \sim \mathbb{P}_{\mathbf{X}}, \mathbf{Y} \sim \mathbb{P}_{\mathbf{Y}}} \left[k_{\text{Sig}} \left(\mathbf{X}, \mathbf{Y}\right) \right].
\end{eqnarray}

\subsection{Empirical Estimator}

To compute the sig-MMD between two distributions, the expectations in Eq.\ \ref{eqn:mmd_kernel} are approximated using empirical samples. Suppose $N$ independent samples 
\begin{equation*}
    \mathbf{X}^{1{:}N} = \left(\mathbf{X}^{1}, \cdots, \mathbf{X}^{N}\right) = \left(\mathbf{X}^{i}\right)_{i=1}^{N} \text{ with } \mathbf{X}^{i} \sim \mathbb{P}_{\mathbf{X}}
\end{equation*}
and $M$ independent samples 
\begin{equation*}
    \mathbf{Y}^{1{:}M} = \left(\mathbf{Y}^{1}, \cdots, \mathbf{Y}^{M}\right) = \left(\mathbf{Y}^{j}\right)_{j=1}^{M} \text{ with } \mathbf{Y}^{j} \sim \mathbb{P}_{\mathbf{Y}}
\end{equation*}
are available. $N, M$ are the sample sizes (batch sizes) and throughout this work, sample size and batch size are used interchangeably. \newline

\noindent
Given empirical samples, the MMD can be computed using either a biased or an unbiased estimator \cite{first_order_mmd_distinguish}. In the numerical examples presented in Section \ref{sec:numerical_examples}, we compare the performance of the two estimators when performing the two-sample hypothesis test. We show that in certain cases, the biased estimator has superior performance over the unbiased estimator. A biased estimator is given by

\begin{eqnarray}\label{Eqn:mmd_biased}
    \nonumber \widehat{d_{k_{\text{Sig}}}^{b}} \left( \mathbb{P}_{\mathbf{X}}, \mathbb{P}_{\mathbf{Y}} \right)^{2} &=& \frac{1}{N^{2}} \sum_{i, j=1}^{N} k_{\text{Sig}} \left(\mathbf{X}^{i}, \mathbf{X}^{j}\right) + \frac{1}{M^{2}} \sum_{i, j=1}^{M} k_{\text{Sig}} \left(\mathbf{Y}^{i}, \mathbf{Y}^{j} \right) \\
    &-& \frac{2}{NM} \sum_{i, j=1}^{N, M} k_{\text{Sig}} \left(\mathbf{X}^{i}, \mathbf{Y}^{j}\right).
\end{eqnarray} 

\noindent
An unbiased estimator is given by
\begin{eqnarray}\label{Eqn:mmd_unbiased}
\nonumber \widehat{d_{k_{\text{Sig}}}} \left( \mathbb{P}_{\mathbf{X}}, \mathbb{P}_{\mathbf{Y}} \right)^{2} &=& \frac{1}{N \left(N-1\right)} \sum_{\substack{i, j=1 \\ i \neq j}}^{N} k_{\text{Sig}} \left(\mathbf{X}^{i}, \mathbf{X}^{j}\right)  + \frac{1}{M \left(M-1\right)} \sum_{\substack{i, j = 1 \\ i \neq j}}^{M} k_{\text{Sig}} \left(\mathbf{Y}^{i}, \mathbf{Y}^{j} \right) \\
&-& \frac{2}{NM} \sum_{i, j=1}^{N, M} k_{\text{Sig}} \left(\mathbf{X}^{i}, \mathbf{Y}^{j}\right).
\end{eqnarray}
\noindent
The bias present in the biased estimator is caused by the cross-terms $k_{\text{Sig}} \left(\mathbf{X}^{i}, \mathbf{X}^{i}\right)$ and $k_{\text{Sig}} \left(\mathbf{Y}^{i}, \mathbf{Y}^{i}\right)$ in the computation of the sample averages.

\subsection{General Signature Kernels}

The authors of \cite{general_sig_kernels, weighted_sig} extend the signature kernel to more generic kernels. A weight function is applied to the level terms of the signature to weight their relative importance in the kernel computation. Let $\phi \colon \mathbb{N} \cup \left\{0 \right\} \mapsto \mathbb{R}^{+}$ be a weight function and $\mathbf{s} = \left(\mathbf{s}_{0}, \mathbf{s}_{1}, \cdots \right)$, $\mathbf{t} = \left(\mathbf{t}_{0}, \mathbf{t}_{1}, \cdots \right) \in \prod_{m \geq 0} \left(\mathbb{R}^{d}\right)^{\otimes m}$. We then define the bilinear form 
\begin{equation*}
    \langle \mathbf{s}, \mathbf{t} \rangle_{\phi} \coloneqq \sum_{m = 0}^{\infty} \phi \left(m\right) \langle \mathbf{s}_{m}, \mathbf{t}_{m} \rangle_{m}
\end{equation*}
\noindent
where 
\begin{equation*}
    \langle \mathbf{s}_{m}, \mathbf{t}_{m} \rangle_{m} \coloneqq \sum_{i_{1}, \cdots, i_{m}=1}^{d} \mathbf{s}_{m}^{i_{1}, \cdots, i_{m}} \mathbf{t}_{m}^{i_{1}, \cdots, i_{m}}. 
\end{equation*}

Given two paths $\mathbf{x}, \mathbf{y} \in \mathcal{X}_{1-\text{Var}}$ and a weight function $\phi \colon \mathbb{N} \cup \left\{0\right\} \mapsto \mathbb{R}^{+}$, the $\phi$-signature kernel is defined as
\begin{equation*}
    k_{\text{Sig}}^{\phi} \left(\mathbf{x}, \mathbf{y}\right) \coloneqq \langle \Phi_{\text{Sig}} \left(\mathbf{x}\right), \Phi_{\text{Sig}} \left(\mathbf{y}\right) \rangle_{\phi}. 
\end{equation*}
\noindent
The following lemma provides a necessary condition on the weight function $\phi$ to guarantee that $\phi$-signature kernel is well-defined \cite[Lemma 2.4]{general_sig_kernels}, \cite[Lemma 2.5]{weighted_sig}. 
\begin{lemma}
    The $\phi$-signature kernel is well-defined provided the function $\phi \colon \mathbb{N} \cup \left\{0 \right\} \mapsto \mathbb{R}^{+}$ is such that the series
    \begin{equation*}
        \sum_{m=0}^{\infty} C^{m} \phi\left(m\right) \left(m!\right)^{-2}
    \end{equation*} is summable for every $C > 0$. 
\end{lemma}
 
Using the $\phi$-signature kernel, a general version of the sig-MMD can be defined. The $\phi$-MMD between the distributions of stochastic processes $\mathbf{X}$, $\mathbf{Y}$ is given by
\begin{eqnarray} \label{eqn:phi_mmd_kernel}
    \nonumber d_{k_{\text{Sig}}^{\phi}} \left(\mathbb{P}_{\mathbf{X}}, \mathbb{P}_{\mathbf{Y}} \right)^{2} &=& \mathbb{E}_{\mathbf{X}, \mathbf{X}^{\prime} \sim \mathbb{P}_{\mathbf{X}}} \left[ k_{\text{Sig}}^{\phi} \left(\mathbf{X}, \mathbf{X}^{\prime}\right) \right]  + \mathbb{E}_{\mathbf{Y}, \mathbf{Y}^{\prime} \sim \mathbb{P}_{\mathbf{Y}}} \left[k_{\text{Sig}}^{\phi} \left(\mathbf{Y}, \mathbf{Y}^{\prime}\right) \right]  \\
    &-& 2 \mathbb{E}_{\mathbf{X} \sim \mathbb{P}_{\mathbf{X}}, \mathbf{Y} \sim \mathbb{P}_{\mathbf{Y}}} \left[k_{\text{Sig}}^{\phi} \left(\mathbf{X}, \mathbf{Y}\right) \right].
\end{eqnarray}
Since the mapping $\mu_{\phi} \colon \mathcal{P} \left(\mathcal{K}\right) \mapsto H_{\phi}$\footnote{$H_{\phi}$ is the RKHS with kernel $k_{\text{Sig}}^{\phi}$.} is injective \cite{weighted_sig}, the $\phi$-MMD is a metric. If $\phi$ is the mapping $\phi \colon \mathbb{N} \cup \left\{0\right\} \mapsto \left\{1\right\}$, the $\phi$-signature kernel is $k_{\text{Sig}}$.

\subsection{Example of General Signature Kernels}

The $\phi$-signature kernel provides a general framework for expanding the set of available signature-based kernels. We now provide a specific example of the $\phi$-signature kernel and its broader implications on the $\phi$-MMD. This $\phi$-MMD is used for the purpose of performing the two-sample hypothesis test in Section \ref{sec:numerical_examples}.

\subsubsection{Scalar Multiplication} \label{sec:scalar_mult}

Let $\mathbf{s} = \left(\mathbf{s}_{0}, \mathbf{s}_{1}, \cdots, \right) \in \prod_{m \geq 0} \left(\mathbb{R}^{d}\right)^{\otimes m}$ and consider the mapping 
\begin{equation*}
    \gamma_{\theta} \colon \prod_{m \geq 0} \left(\mathbb{R}^{d}\right)^{\otimes m} \to \prod_{m \geq 0} \left(\mathbb{R}^{d}\right)^{\otimes m} \text{ defined by } \gamma_{\theta} \left(\mathbf{s}\right) = \sum_{m \geq 0} \theta^{m} \mathbf{s}_{m}
\end{equation*}
\noindent
for some scalar $\theta \in \mathbb{R}^{+}$. Incorporating this within the bilinear form $\langle \cdot, \cdot \rangle_{\phi}$, we have that \cite[Lemma 2.9]{general_sig_kernels}
\begin{equation*}
    \langle \gamma_{\theta} \mathbf{s}, \mathbf{t} \rangle_{\phi \left(m\right) \mapsto 1} = \langle \mathbf{s}, \gamma_{\theta} \mathbf{t} \rangle_{\phi \left(m\right)\mapsto 1} = \langle \mathbf{s}, \mathbf{t} \rangle_{\phi \left(m\right) \mapsto \theta^{m}}. 
\end{equation*}
Applying the homomorphism $\gamma_{\theta}$ to elements of the extended tensor algebra is equivalent to scaling the individual bilinear forms $\langle \cdot, \cdot \rangle_{m}$ within the overall computation of the bilinear form $\langle \cdot, \cdot \rangle_{\phi}$. In our specific context, $\mathbf{s}$ and $\mathbf{t}$ are signatures corresponding to two paths $\mathbf{x}$ and $\mathbf{y}$ respectively. In this case, the following holds \cite[Corollary 2.10]{general_sig_kernels}
\begin{equation*}
    k_{\text{Sig}} \left(\theta \mathbf{x}, \mathbf{y} \right) = k_{\text{Sig}} \left(\mathbf{x}, \theta \mathbf{y} \right) = k_{\text{Sig}} \left(\sqrt{\theta}\mathbf{x}, \sqrt{\theta} \mathbf{y}\right) = k_{\text{Sig}}^{\phi} \left(\mathbf{x}, \mathbf{y}\right)
\end{equation*}
where $\phi$ is the weight function $\phi_{\theta} \colon \mathbb{N} \cup \left\{0\right\} \to \mathbb{R}$ defined by $\phi \left(m\right) = \theta^{m}$. Multiplying paths by a constant scalar $\sqrt{\theta}$ equates to using a $\phi$-signature kernel with weight function $\phi_{\theta}$. Therefore, if both paths are multiplied by the scalar $\sqrt{\theta}$, the inner product $\langle \cdot, \cdot \rangle_{m}$ corresponding to level $m$ is scaled by a factor of $\theta^{m}$. Through path scaling, either one of the following three scenarios holds:
\begin{enumerate}
    \item If $\theta < 1$, a higher weighting is attributed to lower-level terms;
    \item If $\theta = 1$, the original signature kernel is used; or
    \item If $\theta > 1$, a higher weighting is attributed to higher-level terms. 
\end{enumerate}
In terms of $\phi$-MMD, by re-weighting the level terms, we can target the $\phi$-MMD to focus on particular distributional properties. For example, by setting a high value for $\theta$ we would be prioritising higher-order moments over lower-order moments in the computation of the $\phi$-MMD. The opposite effect occurs if we use values for $\theta$ which are less than $1$. A key difference between the $\phi$-MMD and the $\phi$-signature kernel is that in the case of the $\phi$-signature kernel, rescaling one of the paths by $\theta$ is equivalent to scaling both paths by $\sqrt{\theta}$. This is not the case for the $\phi$-MMD. The $\phi$-MMD is a function of kernels of the forms 
\vspace{0.6ex}
\begin{center}
\begin{itemize}[itemjoin=\quad,
  before=\hspace{-1.5ex},
  after=\hspace{1.5ex}]
    \item $k_{\text{Sig}}^{\phi} \left(\mathbf{X}, \mathbf{X}^{\prime}\right)$
    \item  $k_{\text{Sig}}^{\phi} \left(\mathbf{Y}, \mathbf{Y}^{\prime}\right)$
    \item  $k_{\text{Sig}}^{\phi} \left(\mathbf{X}, \mathbf{Y}\right)$
\end{itemize}
\end{center}
\vspace{0.6ex}
\noindent
with inputs being $\mathbb{P}_{\mathbf{X}}, \mathbb{P}_{\mathbf{Y}}$. If we scale $\mathbb{P}_{\mathbf{X}}$ by $\theta$, we would be re-weighting the terms $k_{\text{Sig}}^{\phi} \left(\mathbf{X}, \mathbf{X}^{\prime}\right)$,  $k_{\text{Sig}}^{\phi} \left(\mathbf{X}, \mathbf{Y}\right)$, and $k_{\text{Sig}}^{\phi} \left(\mathbf{Y}, \mathbf{Y}^{\prime}\right)$ by $\theta^{2}$, $\theta$, and $1$ respectively. Hence, inconsistent scalings are being applied and, more importantly, different $\phi$-signature kernels are being used to compute the $\phi$-MMD. We need to scale both inputs and scaling by $\sqrt{\theta}$ is equivalent to re-weighting the signature terms of level $m$ by $\theta^{m}$.

\subsection{Detailed Construction of the $\phi$-MMD}

We reformulate the computation of the $\phi$-MMD as a sum over \textit{level terms} - the signature terms corresponding to a specific level of the signature. We use this reformulation to quantify the contribution of the level terms to the $\phi$-MMD.

\begin{definition} \label{defn:word}
    An alphabet $\mathcal{A}_{d}$ is a set comprising of the $d$ letters $1, \cdots, d$. A word of length $k$ is a sequence of $k$ letters from the alphabet (repetitions are allowed). The set of all words of length $k$ over alphabet $\mathcal{A}_{d}$ is denoted by $\mathcal{W}_{k} \left(\mathcal{A}_{d}\right)$.
\end{definition}

For any two $d$-dimensional stochastic processes $\mathbf{X}, \mathbf{Y}$ and any $m \in \mathbb{N}$, define the function $\Lambda_{m} \colon \mathcal{P} \left(\mathcal{K}\right) \times \mathcal{P} \left(\mathcal{K}\right) \mapsto \mathbb{R}$ by

\begin{equation*}
    \Lambda_{m} \left(\mathbb{P}_{\mathbf{X}}, \mathbb{P}_{\mathbf{Y}} \right) \coloneqq \sum_{i_{1}, \cdots, i_{m} = 1}^{d} \mathbb{E}_{\mathbf{X} \sim \mathbb{P}_{\mathbf{X}}} \left[ \Phi_{\text{Sig}, m} \left(\mathbf{X}\right)^{i_{1}, \cdots, i_{m}} \right] \mathbb{E}_{\mathbf{Y} \sim \mathbb{P}_{\mathbf{Y}}} \left[ \Phi_{\text{Sig}, m} \left(\mathbf{Y}\right)^{i_{1}, \cdots, i_{m}} \right].
\end{equation*}
$\Lambda_{m}$ is the level-$m$ contribution to the expected signature kernel. This can be used to define the level-$m$ contribution to the $\phi$-MMD. The level-$m$ contribution is the mapping $\Gamma_{m}^{\phi} \colon \mathcal{P} \left(\mathcal{K}\right) \times \mathcal{P} \left(\mathcal{K}\right) \mapsto \mathbb{R}$ defined by
\begin{equation*}
    \Gamma_{m}^{\phi} \left(\mathbb{P}_{\mathbf{X}}, \mathbb{P}_{\mathbf{Y}}\right) 
    \coloneqq \phi \left(m\right) \left[ \Lambda_{m} \left(\mathbb{P}_{\mathbf{X}}, \mathbb{P}_{\mathbf{Y}} \right) - 2 \Lambda_{m} \left(\mathbb{P}_{\mathbf{X}}, \mathbb{P}_{\mathbf{Y}} \right) +\Lambda_{m} \left(\mathbb{P}_{\mathbf{X}}, \mathbb{P}_{\mathbf{Y}} \right) \right]. 
\end{equation*}
Using the level-$m$ contribution $\Gamma_{m}^{\phi}$, the $\phi$-MMD can be reformulated as
\begin{equation*}
    d_{k_{\text{Sig}}^{\phi}} \left(\mathbb{P}_{\mathbf{X}}, \mathbb{P}_{\mathbf{Y}} \right)^{2} = \sum_{m \geq 0} \Gamma_{m}^{\phi} \left(\mathbb{P}_{\mathbf{X}}, \mathbb{P}_{\mathbf{Y}}\right).
\end{equation*}
By restricting the sum to the first $k$ levels, we obtain the truncated $\phi$-MMD at level $k$. \newline

Suppose we have $N$ independent and identically distributed (i.i.d.) samples $\mathbf{X}^{1{:}N}$ and $M$ i.i.d.\ samples $\mathbf{Y}^{1{:}M}$. Let $\delta_{\left(\cdot \right)}$ denote the Dirac measure and for two paths $\mathbf{X}, \mathbf{Y} \in \mathcal{K}$ denote $\Lambda_{m} \left( \delta_{\mathbf{X}}, \delta_{\mathbf{Y}} \right)$ by $\Lambda_{m} \left(\mathbf{X}, \mathbf{Y}\right)$. The expectations $\Lambda_{m}$ can be estimated using an unbiased or biased sample average. The unbiased estimators are given by

\begin{eqnarray*}
    \widehat{\Lambda_{m}} \left(\delta_{\mathbf{X}^{1{:}N}}, \delta_{\mathbf{Y}^{1{:}M}} \right) &=& \frac{1}{NM} \sum_{i, j=1}^{N, M} \widehat{\Lambda_{m}} \left(\mathbf{X}^{i}, \mathbf{Y}^{j}\right) \\\nonumber
    &=& \frac{1}{NM} \sum_{i, j=1}^{N, M} \sum_{r_{1}, \cdots, r_{m}=1}^{d} \Phi_{\text{Sig}, m} \left(\mathbf{X}^{i}\right)^{r_{1}, \cdots, r_{m}} \Phi_{\text{Sig}, m} \left(\mathbf{Y}^{j}\right)^{r_{1}, \cdots, r_{m}},  
\end{eqnarray*}
\noindent
and
\begin{eqnarray*}
    &\widehat{\Lambda_{m}}& \left(\delta_{\mathbf{X}^{1{:}N}}, \delta_{\mathbf{X}^{1{:}N}} \right) \\\nonumber
    &=& \frac{1}{N\left(N-1\right)} \sum_{\substack{i, j=1 \\ i \neq j}}^{N} \widehat{\Lambda_{m}} \left(\mathbf{X}^{i}, \mathbf{X}^{j}\right) \\\nonumber
    &=& \frac{1}{N\left(N-1\right)} \sum_{\substack{i, j=1 \\ i \neq j}}^{N} \sum_{r_{1}, \cdots, r_{m}=1}^{d} \Phi_{\text{Sig}, m} \left(\mathbf{X}^{i}\right)^{r_{1}, \cdots, r_{m}}\Phi_{\text{Sig}, m} \left(\mathbf{X}^{j}\right)^{r_{1}, \cdots, r_{m}}. 
\end{eqnarray*}
The biased estimator is denoted by $\widehat{\Lambda_{m}^{b}}$ and is defined by taking the sum over all terms $i, j$ including the cross terms. Using the empirical estimator for the $\Lambda_{m}$ terms, an approximator for the level-$m$ contribution to the $\phi$-MMD is given by
\begin{align*}
    \widehat{\Gamma_{m}^{\phi}} \left( \delta_{\mathbf{X}^{1{:}N}}, \delta_{\mathbf{Y}^{1{:}M}} \right) = \phi \left(m\right) \Big[ &\widehat{\Lambda_{m}} \left(\delta_{\mathbf{X}^{1{:}N}}, \delta_{\mathbf{X}^{1{:}N}} \right) - 2\widehat{\Lambda_{m}} \left(\delta_{\mathbf{X}^{1{:}N}}, \delta_{\mathbf{Y}^{1{:}M}} \right) \\\nonumber
    &~~+ \widehat{\Lambda_{m}} \left(\delta_{\mathbf{Y}^{1{:}M}}, \delta_{\mathbf{Y}^{1{:}M}} \right)\Big].
\end{align*}
Finally, an unbiased estimator for the $\phi$-MMD is
\begin{equation*}
    \widehat{d_{k_{\text{Sig}}^{\phi}}} \left( \delta_{\mathbf{X}^{1{:}N}}, \delta_{\mathbf{Y}^{1{:}M}} \right)^{2} = \sum_{m \geq 0} \widehat{\Gamma_{m}^{\phi}} \left( \delta_{\mathbf{X}^{1{:}N}}, \delta_{\mathbf{Y}^{1{:}M}} \right)
\end{equation*}
with the biased estimator being
\begin{equation*}
    \widehat{d_{k_{\text{Sig}}^{\phi}}^{b}} \left( \delta_{\mathbf{X}^{1{:}N}}, \delta_{\mathbf{Y}^{1{:}M}} \right)^{2} = \sum_{m \geq 0} \widehat{\Gamma_{m}^{\phi, b}} \left( \delta_{\mathbf{X}^{1{:}N}}, \delta_{\mathbf{Y}^{1{:}M}} \right).
\end{equation*}
Whenever the weight function $\phi$ is the constant function $\phi \left(m\right) = 1$ for all $m$, the $\phi$ superscript is omitted from the notation. \newline

By reformulating the MMD in terms of level contributions, we can better understand the role the individual levels play in the computation of the MMD. Since the level-$K$ term of the expected signature is associated with the $K^{\text{th}}$ moment of the distribution of the stochastic process, differences between the $K^{\text{th}}$ moment of the distributions start featuring in the computation of the MMD as from the level-$K$ term $\Gamma_{K}^{\phi}$. Suppose the two distributions have equal moments up to the $\left(K-1\right)^{\text{th}}$ moment and differ in their $K^{\text{th}}$ moment. Since the signature terms decay factorially, the terms $\widehat{\Gamma_{m}}$ are of order $\left(m!\right)^{-2}$ for all $m \leq K$. Hence, terms corresponding to equal moments contribute more to the final value of the MMD and therefore, using the MMD as a statistic on which to perform the two-sample hypothesis test could result in a high probability of a Type 2 error occurring. To counteract the effect of the factorial decay, the $\phi$-MMD can be used to re-weight the level contributions. Generally, by using constant mappings $\phi \colon m \mapsto \theta^{m}$ for some $\theta \in \mathbb{R}^{+}$, higher-order properties can be captured since higher weightings are given to higher level signature terms (Section \ref{sec:scalar_mult}). Numerical examples illustrating this are provided in Section \ref{sec:numerical_examples}. \newline

Constant mappings can be regarded as a change of units. For example, if the sample paths correspond to percentage returns and are provided within a normalised range, changing the units to percentages corresponds to the $\phi$-MMD with $\theta = 100^{2}$. As shown in Section \ref{sec:numerical_examples}, scalings have an effect on test performance. Consequently, the test performed with normalised returns could result in a different conclusion than if returns provided as a percentage were used. In contrast to other test statistics (such as the Kolmogorov-Smirnov test statistic \cite{ks_scale_invariant}), test performance with respect to the sig-MMD is not scale invariant. On the contrary, the scaling factor should be optimised since it plays an important role in test performance.

\section{Two-Sample Hypothesis Testing} \label{sec:type_2_error}

The standard two-sample hypothesis test between stochastic processes \cite{sig_mom_charac_law} is used to test whether two collections of time series stem from the same distribution. The null hypothesis under the two-sample test is
\begin{equation*}
H_{0} \colon \mathbb{P}_{\mathbf{X}} = \mathbb{P}_{\mathbf{Y}}
\end{equation*}
with the alternative hypothesis being
\begin{equation*}
    H_{1} \colon \mathbb{P}_{\mathbf{X}} \neq \mathbb{P}_{\mathbf{Y}}.
\end{equation*}
For the remainder of this section, we fix the distributions $\mathbb{P}_{\mathbf{X}}$ and $\mathbb{P}_{\mathbf{Y}}$. We describe the procedure and the key components required to perform the two-sample test. \newline

Since the MMD is a metric on the space of Borel probability measures on stochastic processes \cite{sig_mom_charac_law, first_order_mmd_distinguish}, it is used as the test statistic for the hypothesis test. Let $\widehat{c}_{1-\alpha}$ be the $\left(1-\alpha\right)$-quantile of the empirical MMD under the null hypothesis. The null hypothesis is rejected with significance $\alpha$ if $\widehat{d_{k_{\text{Sig}}}^{2}} \left(\mathbb{P}_{\mathbf{X}}, \mathbb{P}_{\mathbf{Y}}\right) < \widehat{c}_{1 - \alpha}$. \newline

In two-sample testing, a Type 1 error occurs when the statistical test determines that the two samples have different underlying distributions when the null hypothesis holds. A Type 2 error occurs when the test concludes that the two samples arise from the same stochastic process (distribution) even though this is not the case. \newline

To quantify the likelihood of a Type 2 error occurring, we need to approximate the distribution of the sig-MMD under both the null and alternative hypotheses. Suppose the empirical MMD has distribution $\widehat{F_{H_{0}}}$ under the null hypothesis and $\widehat{F_{H_{1}}}$ under the alternative hypothesis. For a given level $\alpha$, the probability of a Type 2 error is 
\begin{equation}\label{eqn:prob_type_2_error_general}
    \mathbb{P} \left[ \text{Type 2 error} \right] = \widehat{F_{H_{1}}} \left(\widehat{c}_{1-\alpha}\right).
\end{equation} 

To perform the two-sample hypothesis test and compute the probabilities of a Type 1 error and Type 2 error, the following are needed:
\begin{enumerate}
    \item An estimate for the $\left(1-\alpha\right)$-quantile of the null distribution; and
    \item The distribution of the MMD under the alternative hypothesis. 
\end{enumerate} 
  
We describe two techniques for approximating the distributions; bootstrapping and asymptotic analysis. The former is based on re-sampling from the stochastic processes until an `accurate'\footnote{Generally, this is a function of the number of samples. However, if large sample sizes are available, it may not be computationally feasible to use all samples.} approximation of the MMD distribution can be established. The latter approach relies heavily on U-statistics \cite{u_stat_4, u_stat_6, u_stat_1, u_stat_5, u_stat_3, u_stat_2}. 

\subsection{Bootstrapping}
Let $\mathbf{X}^{1{:}N^{\prime}}$ and $\mathbf{Y}^{1{:}M^{\prime}}$ be two independent collections of paths sampled from $\mathbb{P}_{X}$ and $\mathbb{P}_{Y}$ respectively with values in $\mathbb{R}^{d}$. To sample the MMD under the null distribution, we sample batches of size $N_{1}, N_{2}$ from the $N^{\prime}$ available. This procedure is repeated $B$ times to construct the collection 
\begin{equation*}
    \left\{\left(\mathbf{X}_{i}^{1{:}N_{1}}, \mathbf{X}_{i}^{1{:}N_{2}} \right)\right\}_{i=1}^{B}.
\end{equation*}
\noindent
The empirical MMD between these samples is calculated to obtain $B$ distances under the null hypothesis. These $B$ distances are used to construct the empirical null distribution. Repeating a similar procedure, the distribution of the test statistic under the alternative hypothesis is approximated by sampling $B$ collections $\left\{\left(\mathbf{X}^{1{:}N_{1}}, \mathbf{Y}^{1{:}M_{1}}\right)\right\}_{i=1}^{B}$ and computing the test statistic over every collection. The value of $B$ is closely related to the quality of the approximating distribution.\newline

Permutation testing \cite{sig_mom_charac_law} is an alternative approach to sampling the test statistic under the null hypothesis. To perform such test, the sample paths $\mathbf{X}^{1{:}N^{\prime}}$ and $\mathbf{Y}^{1{:}M^{\prime}}$ are pooled together. Sampling the empirical MMD under the null is performed by computing the empirical MMD between two separate collections of size $N^{\prime}, M^{\prime}$ respectively. These collections are sampled uniformly from all possible permutations of the pooled dataset into two groups of size $N^{\prime}$ and $M^{\prime}$. 

\subsection{U-Statistics}

Let $\mathcal{M}$ be a compact metric space and consider a distribution $\mu \in \mathcal{P} \left(\mathcal{M}\right)$. Suppose we have samples $x_{1}, \cdots, x_{N} \sim \mu$. Let $h \colon \mathcal{M}^{r} \to \mathbb{R}$ be a kernel function with $r \leq N$. The U-statistic \cite{u_stat_4, u_stat_6, u_stat_1, u_stat_5, u_stat_3, u_stat_2} with kernel $h$ is defined as
\begin{equation*}
    U_{N} \coloneqq \frac{\left(N-r\right)!}{N!} \sum_{\left(i_{1}, \cdots, i_{r}\right) \in \mathbf{P}_{r, N}} h \left(x_{i_{1}}, \cdots, x_{i_{r}}\right)
\end{equation*}
where $\mathbf{P}_{r, N}$ denotes the set of all $N!/\left(N-r\right)!$ permutations of size $r$ chosen from $\left\{1, \cdots, N\right\}$. If $h$ is symmetric, then
\begin{equation*}
    U_{N} = \frac{1}{\binom{N}{r}} \sum_{\left(i_{1}, \cdots, i_{r}\right) \in \mathbf{C}_{r, N}} h \left(x_{i_{1}}, \cdots, x_{i_{r}}\right)
\end{equation*}
where $\mathbf{C}_{r, N}$ denotes the set of all combinations of $r$ integers from $\left\{1, \cdots, N\right\}$ such that $i_{1} \leq \cdots \leq i_{r}$. For example, if $r=1$, the sample mean $N^{-1} \sum_{i} x_{i}$ is a U-statistic. By setting $h$ to be a linear combination of $\phi$-signature kernels, asymptotic results regarding limiting distributions of U-statistics can be applied to the $\phi$-MMD. 

\subsection{Asymptotic Analysis: Null Distribution}

To perform the two-sample hypothesis test and quantify the probabilities of Type 1 and Type 2 errors occurring, an estimate for the threshold value $\widehat{c}_{1-\alpha}$ is needed. Since $\widehat{c}_{1-\alpha}$ is the $\alpha$-quantile of the null distribution $\widehat{F_{H_{0}}}$, by approximating the null distribution, the $\left(1-\alpha\right)$-quantile can then be computed. \newline

Suppose we have $N$ samples $\mathbf{X}^{1{:}N}, \mathbf{Y}^{1{:}N}$. Define $\mathbf{Z}^{i} \coloneqq \left(\mathbf{X}^{i}, \mathbf{Y}^{i}\right)$. Another unbiased estimator for $d_{k_{\text{Sig}}} \left( \mathbb{P}_{\mathbf{X}}, \mathbb{P}_{\mathbf{Y}} \right)^{2}$ is given by the one sample U-statistic \cite[Lemma 7]{mmd_null}
\begin{equation*}
    \widehat{d_{k_{\text{Sig}}, 2}} \left(\mathbb{P}_{\mathbf{X}}, \mathbb{P}_{\mathbf{Y}}\right)^{2} = \frac{1}{N \left(N-1\right)} \sum_{\substack{i, j=1 \\ i \neq j}}^{N} h \left(\mathbf{Z}^{i}, \mathbf{Z}^{j}\right)
\end{equation*}
where $h \left(\cdot, \cdot\right)$ is defined by
\begin{equation*}
    h \left(\mathbf{Z}^{i}, \mathbf{Z}^{j}\right) \coloneqq k_{\text{Sig}} \left(\mathbf{X}^{i}, \mathbf{X}^{j} \right) + k_{\text{Sig}} \left(\mathbf{Y}^{i}, \mathbf{Y}^{j}\right) - k_{\text{Sig}} \left(\mathbf{X}^{i}, \mathbf{Y}^{j}\right) - k_{\text{Sig}} \left(\mathbf{X}^{j}, \mathbf{Y}^{i}\right). 
\end{equation*}
The difference between $\widehat{d_{k_{\text{Sig}}, 2}}$ and $\widehat{d_{k_{\text{Sig}}}}$ (Eq.\ \eqref{Eqn:mmd_unbiased}) is that the terms $k \left(\mathbf{X}^{i}, \mathbf{Y}^{i}\right)$ are not included in the computation of $\widehat{d_{k_{\text{Sig}}, 2}}$ whilst they are accounted for when computing $\widehat{d_{k_{\text{Sig}}}}$. As described by the authors of \cite{mmd_null}, assuming $\mathbb{E} \left[ h^{2}\right] < \infty$, when applied to larger sample sizes, the null distribution can be approximated by an infinite weighted sum of shifted independent $\chi_{1}^{2}$ random variables as follows
\begin{equation*}
    \frac{1}{N} \widehat{d_{k_{\text{Sig}}, 2}}\left( \mathbb{P}_{\mathbf{X}}, \mathbb{P}_{\mathbf{Y}} \right)^{2} \sim \sum_{i=1}^{\infty} \lambda_{i} \left[z_{i}^{2} - 2 \right]
\end{equation*}
where $z_{i}$ are i.i.d.\ mean-zero normally distributed random variables with variance $2$. The weight $\lambda_{l}$ is the solution to the eigenvalue problem 
\begin{equation*}
    \int_{\mathcal{K}} \tilde{k} \left(\tilde{\mathbf{X}}, \mathbf{X}^{\prime}\right) \psi_{l} \left( \tilde{\mathbf{X}} \right) d \mathbb{P}_{\mathbf{X}} \left(\tilde{\mathbf{X}}\right) = \lambda_{l} \psi_{l} \left(\mathbf{X}^{\prime}\right)
\end{equation*}
where 
\begin{eqnarray*}
    \tilde{k} \left(\tilde{\mathbf{X}}, \mathbf{X}^{\prime}\right) &\coloneqq& k_{\text{Sig}} \left(\tilde{\mathbf{X}}, \mathbf{X}^{\prime}\right) - \mathbb{E}_{\mathbf{X}_{1}} \left[ k_{\text{Sig}} \left(\tilde{\mathbf{X}}, \mathbf{X}_{1}\right) \right] \\\nonumber
    &-& \mathbb{E}_{\mathbf{X}_{2}} \left[ k_{\text{Sig}} \left(\mathbf{X}_{2}, \mathbf{X}^{\prime}\right) \right] + \mathbb{E}_{\mathbf{X}_{1}, \mathbf{X}_{2}} \left[ k_{\text{Sig}} \left(\mathbf{X}_{1}, \mathbf{X}_{2}\right) \right].
\end{eqnarray*}
This approach does not necessitate the same number of samples from the two distributions and it can be computationally intensive as it is based on matrix computations \cite{sig_mmd_example}. Alternatively, the null distribution of the biased MMD can be approximated using two different techniques, both based on low-order moments of the empirical MMD. One of these approaches uses Pearson curves \cite{mmd_null, first_order_mmd_distinguish, fast_consistent_two_sample_test}. The other approach is computationally more efficient and is based on approximating the null distribution of the biased MMD using a gamma distribution \cite{fast_consistent_two_sample_test}. The authors of \cite{mmd_gamma_book} show that
\begin{equation*}
    N\widehat{d_{k_{\text{Sig}}}^{b}} \left(\mathbb{P}_{\mathbf{X}}, \mathbb{P}_{\mathbf{Y}}\right)^{2} \sim \frac{x^{\tau-1}e^{-x/\psi}}{\psi^{\tau} \Gamma \left(\tau\right)}
\end{equation*}
where
$\Gamma \left(\cdot \right)$ is the gamma function and
\begin{equation*}
    \tau \coloneqq \frac{\mathbb{E} \left[ \widehat{d_{k_{\text{Sig}}}^{b}} \left(\mathbb{P}_{\mathbf{X}}, \mathbb{P}_{\mathbf{Y}}\right)^{2} \right]^{2}}{\text{Var} \left( \widehat{d_{k_{\text{Sig}}}^{b}} \left(\mathbb{P}_{\mathbf{X}}, \mathbb{P}_{\mathbf{Y}}\right)^{2} \right)}~~\text{and}~~ \psi \coloneqq \frac{N \text{Var} \left( \widehat{d_{k_{\text{Sig}}}^{b}} \left(\mathbb{P}_{\mathbf{X}}, \mathbb{P}_{\mathbf{Y}}\right)^{2} \right)}{\mathbb{E} \left[ \widehat{d_{k_{\text{Sig}}}^{b}} \left(\mathbb{P}_{\mathbf{X}}, \mathbb{P}_{\mathbf{Y}}\right)^{2} \right]}. 
\end{equation*}
When closed-form solutions are available for truncated estimates of the two moments, these closed-form solutions are used. When these are not available, empirical estimates are constructed using bootstrapping.

\subsection{Asymptotic Analysis: Alternative Distribution}

If $\mathbb{E} \left[ h^{2}\right] < \infty$, the distribution of the squared MMD under the alternative hypothesis converges in distribution to a Gaussian according to \cite[Theorem 8]{mmd_null}
\begin{equation*}
    \sqrt{N} \left(\widehat{d_{k_{\text{Sig}}, 2}}\left( \mathbb{P}_{\mathbf{X}}, \mathbb{P}_{\mathbf{Y}} \right)^{2} - d_{k_{\text{Sig}}}\left(\mathbb{P}_{\mathbf{X}}, \mathbb{P}_{\mathbf{Y}} \right)^{2} \right) \to \mathcal{N} \left(0, \sigma^{2}\right)
\end{equation*}
where 
\begin{equation*}
    \sigma^{2} \coloneqq 4 \left( \mathbb{E}_{\mathbf{Z}} \left[ \mathbb{E}_{\mathbf{Z}^{\prime}} \left[h \left(\mathbf{Z}, \mathbf{Z}^{\prime} \right) \right]^{2}\right] - \mathbb{E}_{\mathbf{Z}, \mathbf{Z}^{\prime}} \left[ h \left(\mathbf{Z}, \mathbf{Z}^{\prime}\right) \right]^{2}\right)
\end{equation*}
and $\mathcal{N} \left(0, \sigma^{2}\right)$ denotes a Gaussian random variable with mean $0$ and variance $\sigma^{2} > 0$.

\section{Numerical Examples} \label{sec:numerical_examples}

In this section we carry out numerical simulations to show the effect different signature kernels have on reducing errors when performing two-sample hypothesis testing. In certain scenarios, the original signature kernel and standard two-sample hypothesis test perform well. In our simulations, we use parametric models to construct examples in which the probability of Type 2 errors occurring are high. This was done to illustrate the importance of re-weighting higher signature levels. The probabilities of a Type 2 error occurring were computed by approximating the distributions of the test statistic under the null and alternative hypotheses using bootstrapping. To compute the probabilities of a Type 1 error occurring, the same procedure as for Type 2 errors was applied, however in this case, both collections of paths were simulated from a stochastic model with the same parameters. When the parameters are chosen such that they are very close, very few $\phi$-signature kernels would reduce the probabilities of errors and more adhoc weight functions may be required. In some situations more sample paths will be needed. In practice, linearly interpolated sample paths are used instead of Brownian motion paths. Although Brownian motion paths do not belong to the set $\mathcal{K}$, the linearly interpolated paths belong to the set $\mathcal{K}$.  Also, the sample paths converge to Brownian motion sample paths (Donsker's invariance principle \cite{donsker}) as the mesh size tends to $0$. Therefore, the linearly interpolated paths adhere to the framework described in Section \ref{sec:fact_decay}. \newline

Throughout the experiments, the significance level was set to $\alpha = 0.05$. Simulations were mainly run on a GPU\footnote{GPUs were provided by the King's College London CREATE environment \cite{create_hpc}.} for computational efficiency. The code used to generate these results can be found at the following repository: \url{https://github.com/Andrew-Alden/SignatureMMDTesting}.

\subsection{Scaled Brownian Motion} \label{sec:brownian_motion_example}

Consider the time interval $\left[0, T\right]$ and two scaled Brownian motions $\mathbf{S}$ and $\mathbf{G}$ with dynamics described by the stochastic differential equations (SDEs)
\begin{equation*}
    d\mathbf{S}_{t} = \sigma dW_{t}^{1}~~\text{and}~~d\mathbf{G}_{t} = \beta dW_{t}^{2}.
\end{equation*}
   
The parameters $\sigma, \beta > 0$ control the volatility of the processes and $\left(W_{t}^{1}\right), \left(W_{t}^{2}\right)$ are independent Brownian motions\footnote{For further details on Brownian motions and their expected signature see \cite{expected_sig_bm_1, expected_sig_bm_2}.}. Define the time-augmented\footnote{Adding the time component removes the possibility of tree-like path occurrences.} processes $\mathbf{X}, \mathbf{Y}$ as $\mathbf{X}_{t} \coloneqq \left(t, \mathbf{S}_{t}\right)$ and $\mathbf{Y}_{t} \coloneqq \left(t, \mathbf{G}_{t}\right)$. The two-sample hypothesis test is performed on the stochastic processes $\mathbf{X}$ and $\mathbf{Y}$. \newline

The processes $\mathbf{S}$ and $\mathbf{G}$ are simulated using the Euler discretisation scheme
\begin{equation*}
    \mathbf{S}_{t_{i}} = \mathbf{S}_{t_{i-1}} + \sigma \sqrt{h} Z, ~~Z \sim \mathcal{N} \left(0, 1\right),~~t_{i} - t_{i-1} = h. 
\end{equation*}
\noindent
It is assumed\footnote{This does not affect the results. If $\mathbf{S}_{0}, \mathbf{G}_{0} \neq 0$ consider the processes $\mathbf{S}^{\prime} \coloneqq \mathbf{S} - \mathbf{S}_{0}$ and $\mathbf{G}^{\prime} \coloneqq \mathbf{G} - \mathbf{G}_{0}$.} that $\mathbf{S}_{0} = \mathbf{G}_{0} = 0$ and we set $T=1$. Since these processes are driftless, 
\begin{equation*}
    \mathbb{E} \left[ \mathbf{S}_{t}\right] = \mathbb{E} \left[ \mathbf{G}_{t} \right] = 0
\end{equation*}
for all $t \in \left[0, T\right]$. If $\sigma \neq \beta$, the processes $\mathbf{S}$ and $\mathbf{G}$ differ from their second moment\footnote{$\mathbb{E} \left[\mathbf{S}_{t}^{2}\right] \neq \mathbb{E} \left[ \mathbf{G}_{t}^{2} \right]$ for all $t \in \left[0, T\right]$.}. Therefore, the sig-MMD should distinguish between these processes. The empirical probabilities of a Type 2 error and a Type 1 error occurring were computed using the empirical distributions of the sig-MMD. To compute probabilities of Type 1 errors occurring, both collections of sample paths were sampled from $\mathbf{X}$ (i.e.\ $H_{0}$). \newline

\begin{figure}[H]
     \centering
     \begin{subfigure}{0.45\textwidth}
         \centering
         \includegraphics[width=0.9\linewidth, height=0.2\textheight]{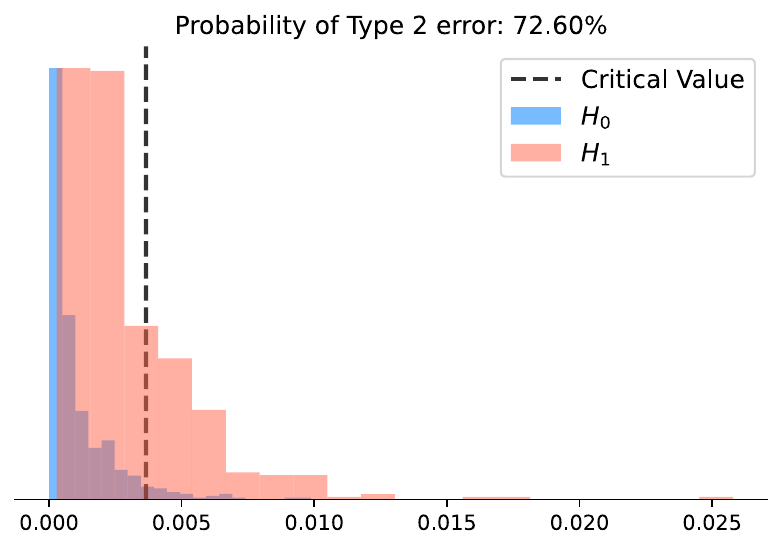}
         \caption{Biased.}
         \label{fig:bm_mmd_biased}
     \end{subfigure}
    \begin{subfigure}{0.45\textwidth}
         \centering
         \includegraphics[width=0.9\linewidth, height=0.2\textheight]{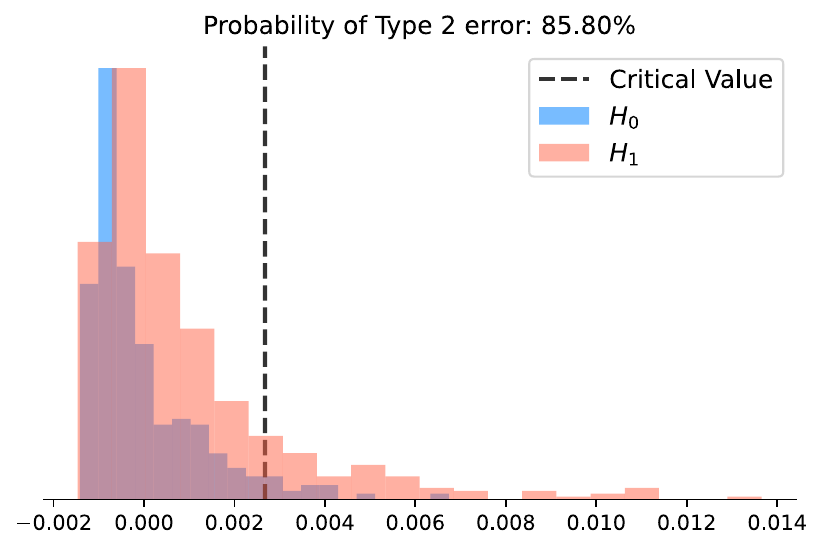}
         \caption{Unbiased.}
         \label{fig:bm_mmd_unbiased}
     \end{subfigure}
          \caption{Null and alternative distributions of the (squared) sig-MMD between two scaled Brownian motions. Batch size of $128$ was used and $500$ independent simulations were run.}
     \label{fig:bm_mmd}
\end{figure}

\begin{figure}[H]
     \centering
     \includegraphics[width=1.0\linewidth, height=0.3\textheight]{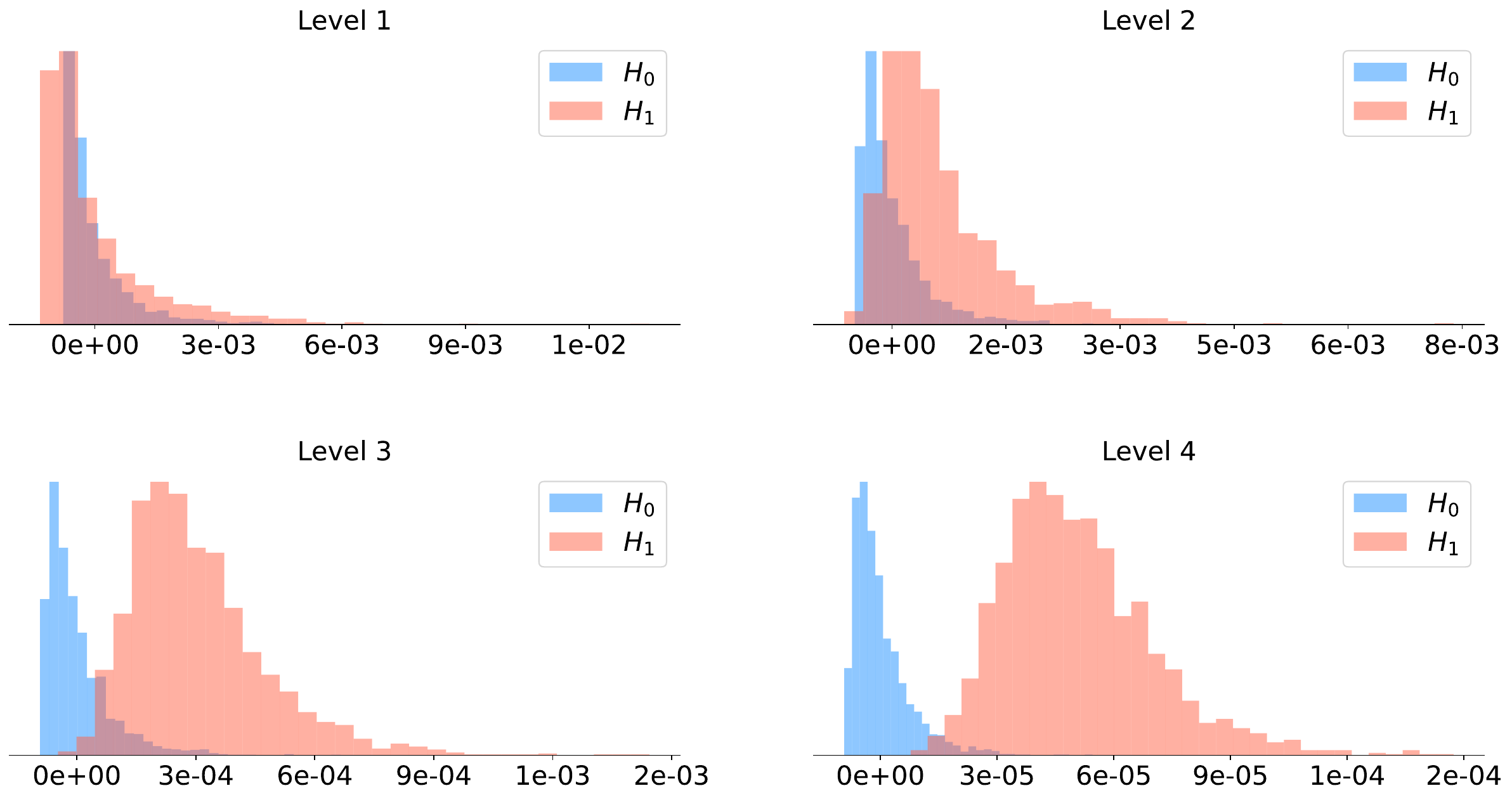}
     \caption{Level contributions to the sig-MMD between two scaled Brownian motions. Batch size of $128$ was used and $2{,}048$ independent simulations were run. No scaling was applied and the unbiased estimator was used.}
     \label{fig:bm_level_contrib}
\end{figure}

In the simulations, $\sigma$ was set to $0.2$, $\beta$ was set to $0.3$, and the batch size was set to $128$. The probability of a Type 2 error occurring was $72.6\%$ with the biased estimator and $85.8\%$ with the unbiased estimator. These probabilities correspond to the large overlap in the histograms in Fig.\ \ref{fig:bm_mmd}. To better understand the underlying cause of the high probability $85.5\%$, the level contributions ($\Gamma_{m}^{\phi}$) were plotted. As can be noted in Fig.\ \ref{fig:bm_level_contrib}, the area of overlap decreases as the signature level increases. As can be seen in the plot of the level-$1$ contribution ($\Gamma_{1}^{\phi}$), the histograms completely overlap. This occurs because the processes are driftless. However, from the second level onwards, the histograms start to gradually separate (region of overlap decreases). Due to the factorial decay of the absolute value of the signature terms, even though the histograms start to separate, the level with the largest contribution to the MMD (first level) corresponds to the plot in which the histograms completely overlap. As a result, the probability of a Type 2 error occurring is high. \newline

By using general signature kernels, the test statistic is tailored to target specific moment(s) when performing the two-sample hypothesis test. In this particular example, scaling the paths allows us to focus on the second moment of the processes. In our simulations, we tested various scaling factors. Scaling was applied to the processes $\mathbf{S}$, $\mathbf{G}$ and not to the time-augmented processes. Therefore, the time-index still ran from $\left[0, T\right]$. This was done because the time-index was used to add a time structure to the process. We could have set $T$ in the time-augmented process to be the inverse of the scaling factor. In this case, multiplying the entire process including the time-index by the scaling factor would result in time starting at $0$ and ending at $T=1$. This is equivalent to our approach of not scaling the time index. \newline

\begin{figure}[H]
     \centering
     \begin{subfigure}{0.45\textwidth}
         \centering
         \includegraphics[width=0.9\linewidth, height=0.2\textheight]{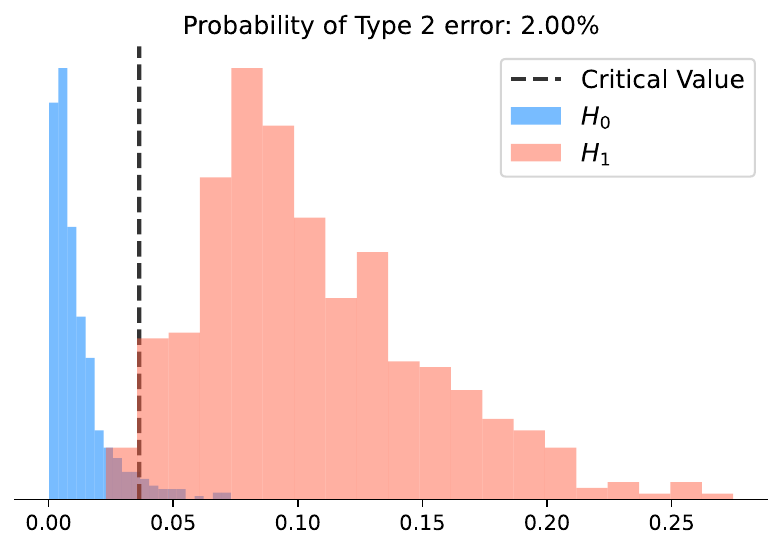}
         \caption{Biased.}
         \label{fig:scaled_bm_mmd_biased}
     \end{subfigure}
    \begin{subfigure}{0.45\textwidth}
         \centering
         \includegraphics[width=0.9\linewidth, height=0.2\textheight]{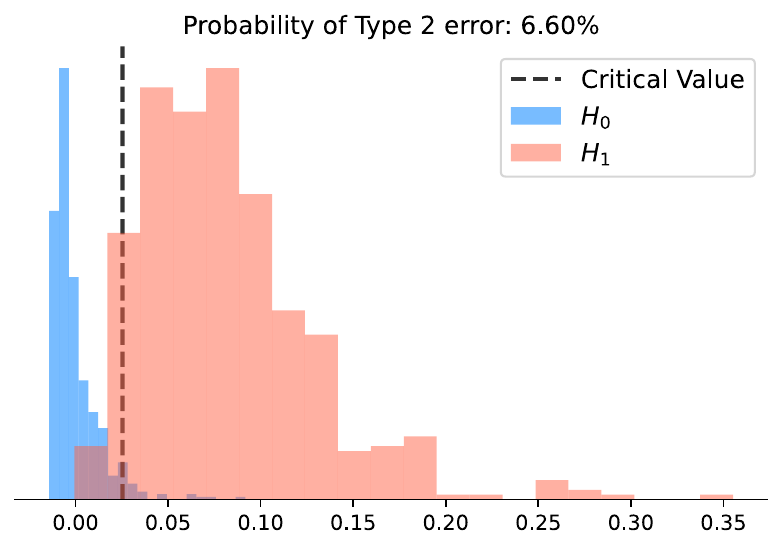}
         \caption{Unbiased.}
         \label{fig:scaled_bm_mmd_unbiased}
     \end{subfigure}
          \caption{Null and alternative distributions of the (squared) $\phi$-MMD between two scaled Brownian motions. Batch size of $128$ was used and $500$ independent simulations were run. A scaling of $3$ was applied.}
     \label{fig:scaled_bm_mmd}
\end{figure}

\begin{figure}[H]
     \centering
     \includegraphics[width=1.0\linewidth, height=0.3\textheight]{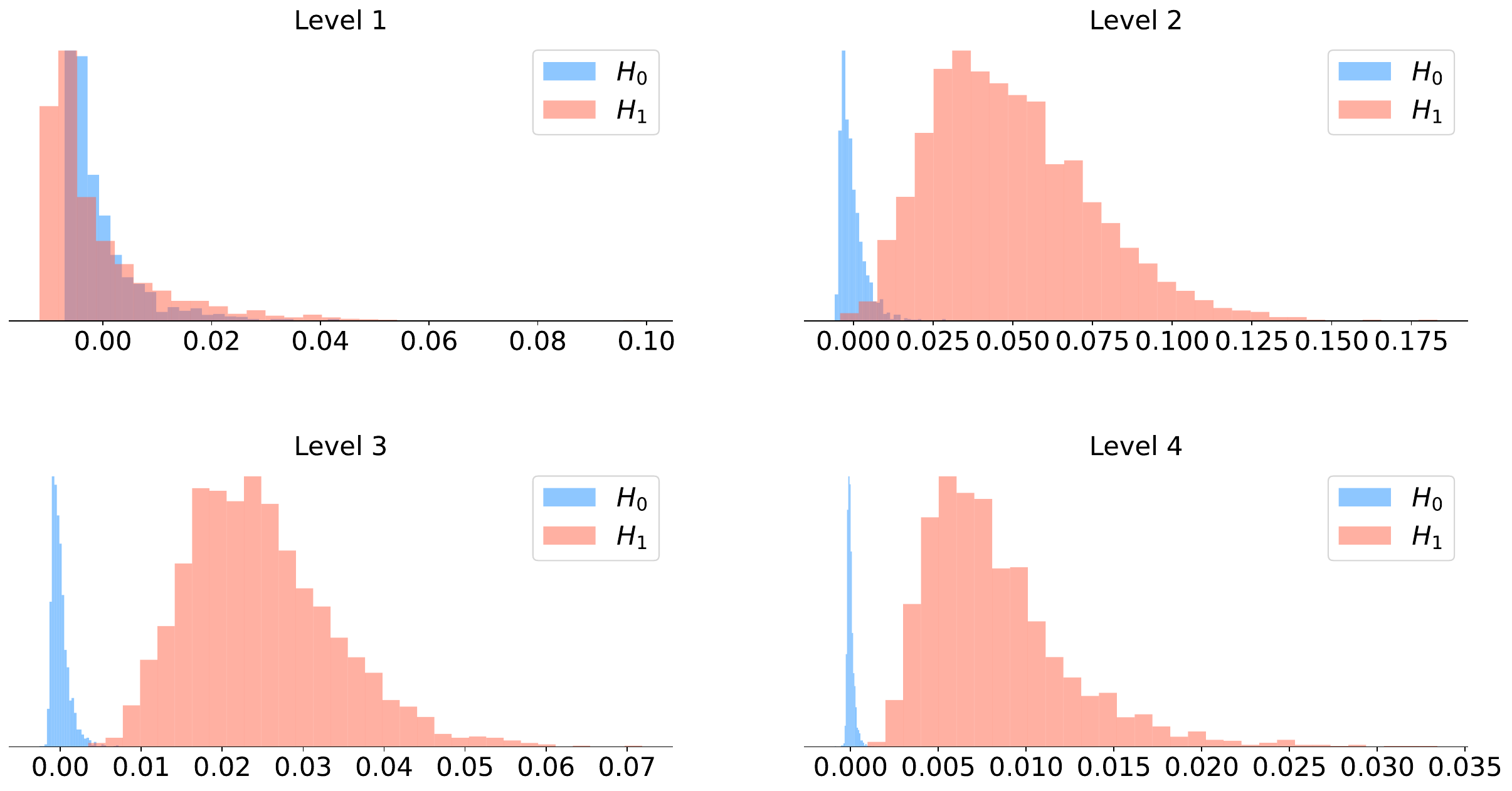}
     \caption{Level contributions to the $\phi$-MMD between two scaled Brownian motions. Batch size of $128$ was used and $2{,}048$ independent simulations were run. A scaling of $3$ was applied and the unbiased estimator was used.}
     \label{fig:scaled_bm_level_contrib}
\end{figure}

When re-running all simulations with scaled paths using a scaling factor of $3$, the probability of a Type 2 error occurring drops to $2.0\%$ (Fig.\ \ref{fig:scaled_bm_mmd_biased}) when using the biased estimator and $6.6\%$ (Fig.\ \ref{fig:scaled_bm_mmd_unbiased}) when using the unbiased estimator. Scaling the paths by a factor of $3$ is equivalent to scaling the parameters $\sigma, \beta$ by a factor of $3$. The scaling amplifies the differences between distributions. The level contributions to the $\phi$-MMD are plotted in Fig.\ \ref{fig:scaled_bm_level_contrib}. The separation of histograms is evident as the level increases. A crucial difference between the histograms in Fig.\ \ref{fig:bm_level_contrib} and Fig.\ \ref{fig:scaled_bm_level_contrib} is that, when scaling is applied (Fig.\ \ref{fig:scaled_bm_level_contrib}), the absolute value of the contribution is larger for the scaled version than for the standard sig-MMD. Hence, higher-order terms which better capture distributional differences contribute more to the final test statistic than when using the standard sig-MMD. \newline 

\begin{figure}[H]
    \centering
    \begin{subfigure}{0.45 \textwidth}
        \includegraphics[width=1.0\linewidth, height=0.2\textheight]{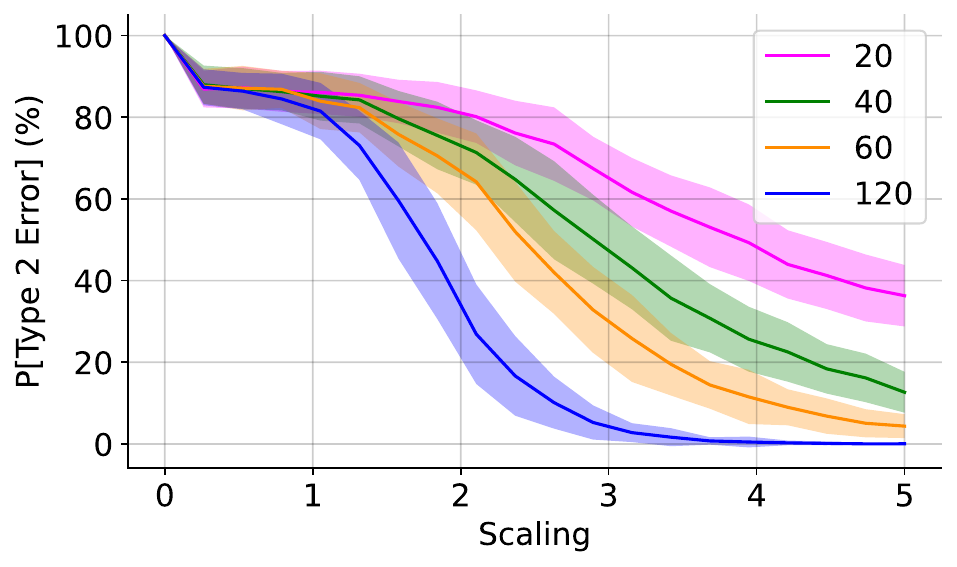}
        \caption{Biased}
        \label{fig:scaling_bm_list_biased}
    \end{subfigure}
        \begin{subfigure}{0.45 \textwidth}
        \includegraphics[width=1.0\linewidth, height=0.2\textheight]{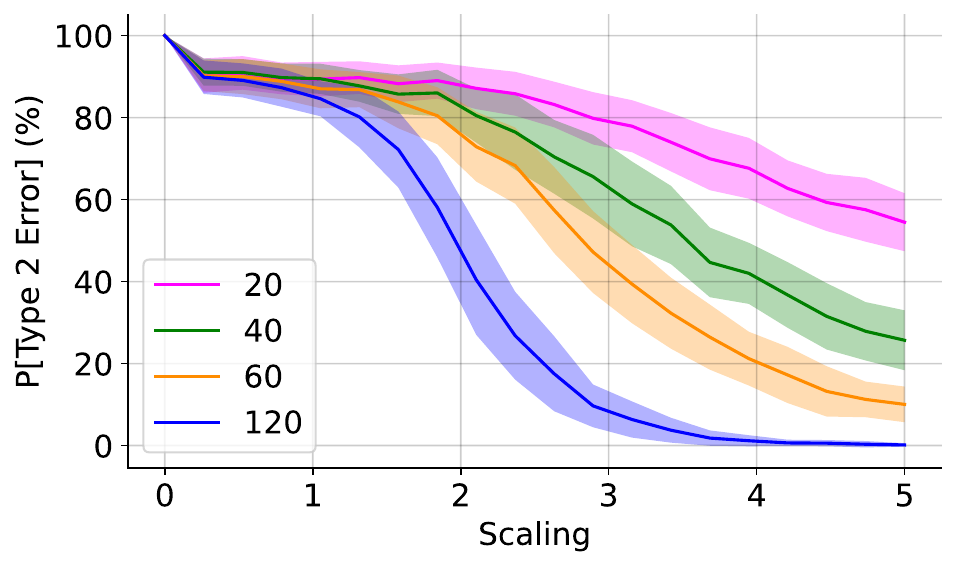}
        \caption{Unbiased}
        \label{fig:scaling_bm_list_unbiased}
    \end{subfigure}

    \begin{subfigure}{0.45 \textwidth}
        \includegraphics[width=1.0\linewidth, height=0.2\textheight]{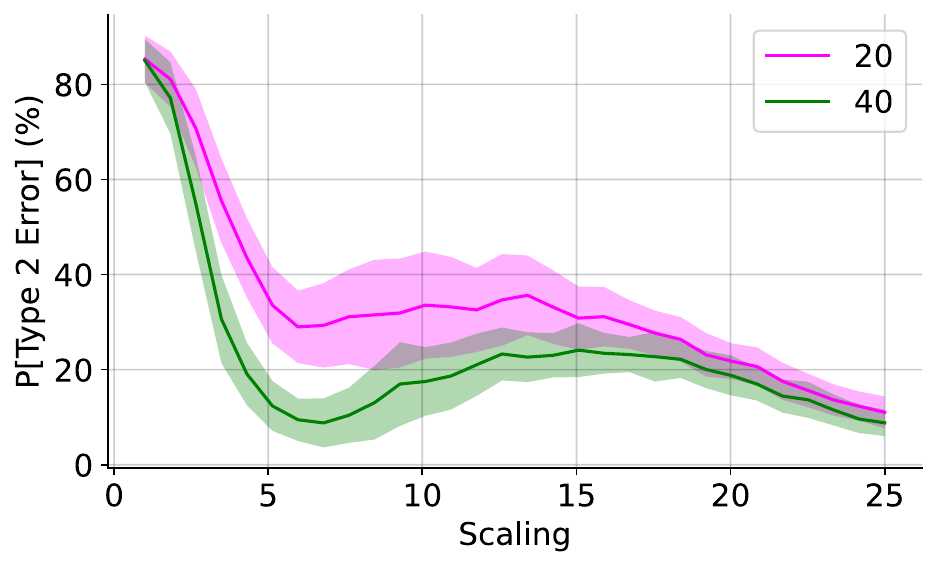}
        \caption{Biased - Extreme Scalings}
        \label{fig:scaling_bm_list_biased_extreme}
    \end{subfigure}
        \begin{subfigure}{0.45 \textwidth}
        \includegraphics[width=1.0\linewidth, height=0.2\textheight]{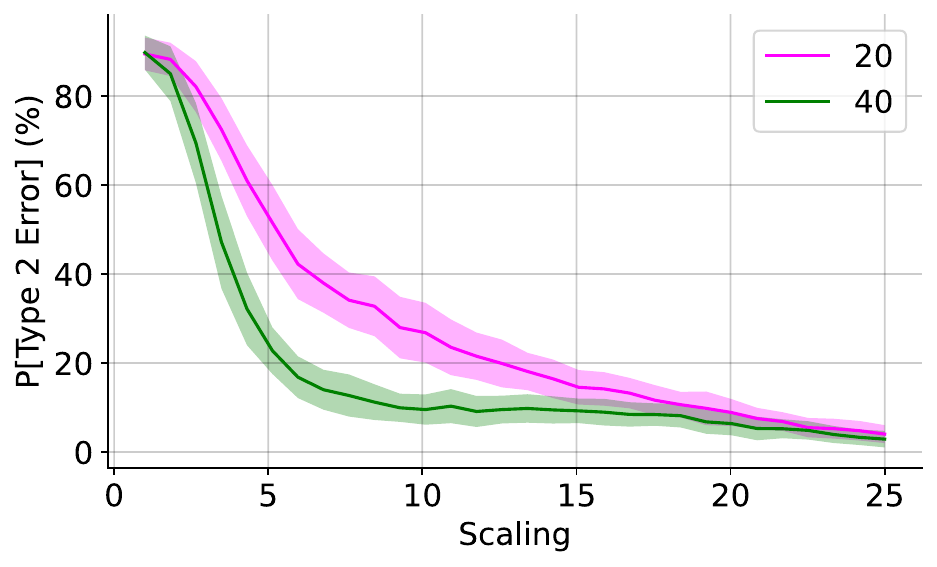}
        \caption{Unbiased - Extreme Scalings}
        \label{fig:scaling_bm_list_unbiased_extreme}
    \end{subfigure}
    
    \caption{Probability of a Type 2 error occurring as a function of scaling factor and batch size between two scaled Brownian motions. Solid line corresponds to the mean and the shaded region corresponds to one standard deviation away from the mean.}
    \label{fig:scaling_bm_list}
\end{figure}

In the above example, the batch size was kept fixed at $128$. It is important to study the effect of scaling paths on the probability of a Type 2 error occurring as a function of the batch size. In Fig.\ \ref{fig:scaling_bm_list}, the probability of a Type 2 error occurring is plotted as a function of the batch size and the scaling factor.  Fig.\ \ref{fig:scaling_bm_list_biased} corresponds to the biased estimator for the $\phi$-MMD and Fig.\ \ref{fig:scaling_bm_list_unbiased} corresponds to the unbiased estimator. For low batch sizes, scaling reduces the probability of a Type 2 error occurring. However, it does not reduce it to levels below $30\%$ unless extreme scalings are used (Fig.\ \ref{fig:scaling_bm_list_biased_extreme} and Fig.\ \ref{fig:scaling_bm_list_unbiased_extreme}). Also, although the impact of scaling is consistent across the biased and the unbiased estimator, the effect takes place earlier in the biased version for scaling values less than $5$. Therefore, for a given batch size, the probability of a Type 2 error occurring is most likely lower when using the biased estimator as opposed to the unbiased estimator if no extreme scaling is applied. \newline

\begin{figure}[H]
    \centering

    \begin{subfigure}{0.95 \textwidth}
    \includegraphics[width=1.0\textwidth, height=0.35\textheight]{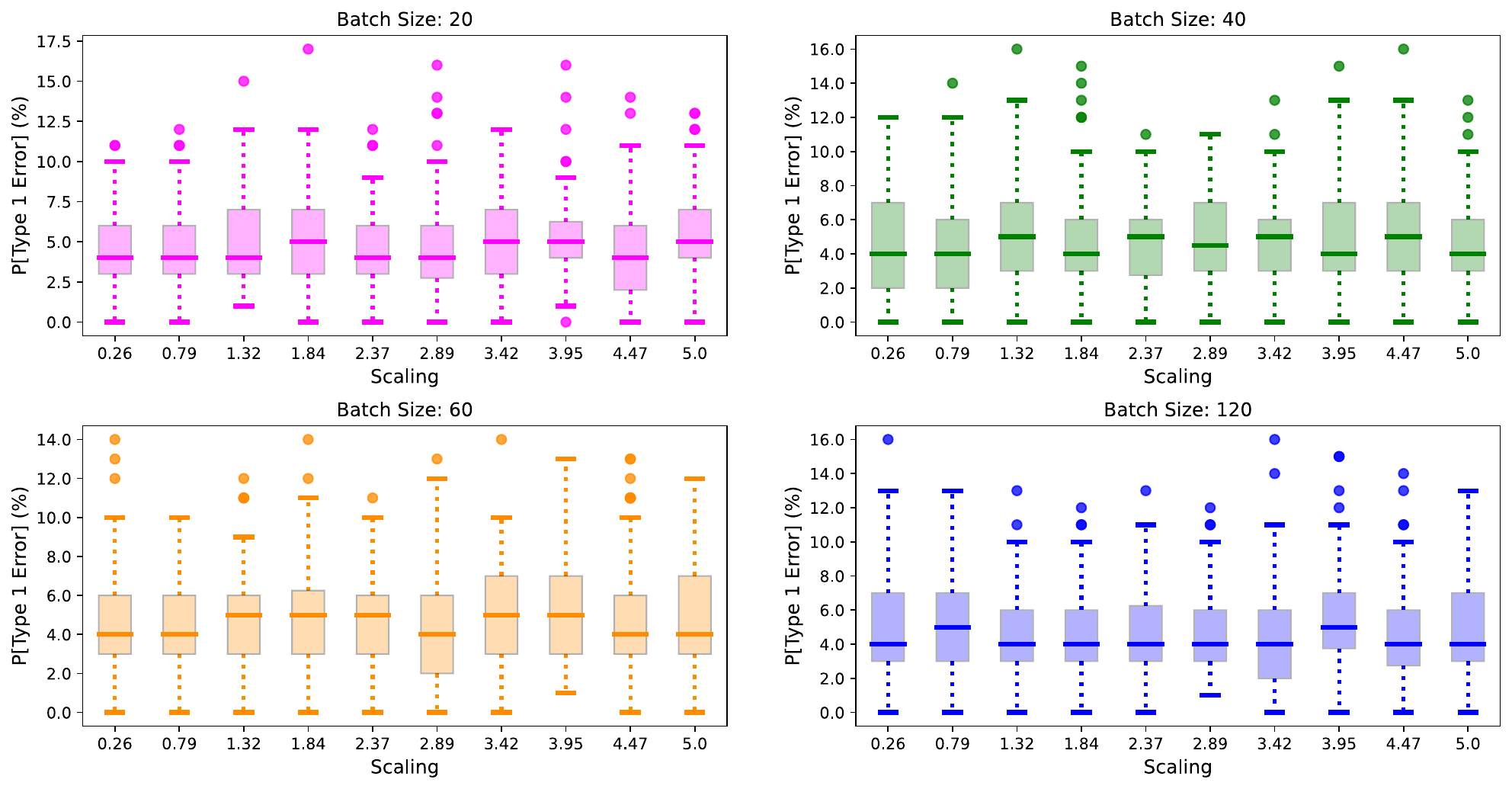}
    \caption{Biased}
    \label{fig:scaling_bm_list_type1_scalings_biased}
    \end{subfigure}
    \begin{subfigure}{0.95 \textwidth}
    \includegraphics[width=1.0\textwidth, height=0.35\textheight]{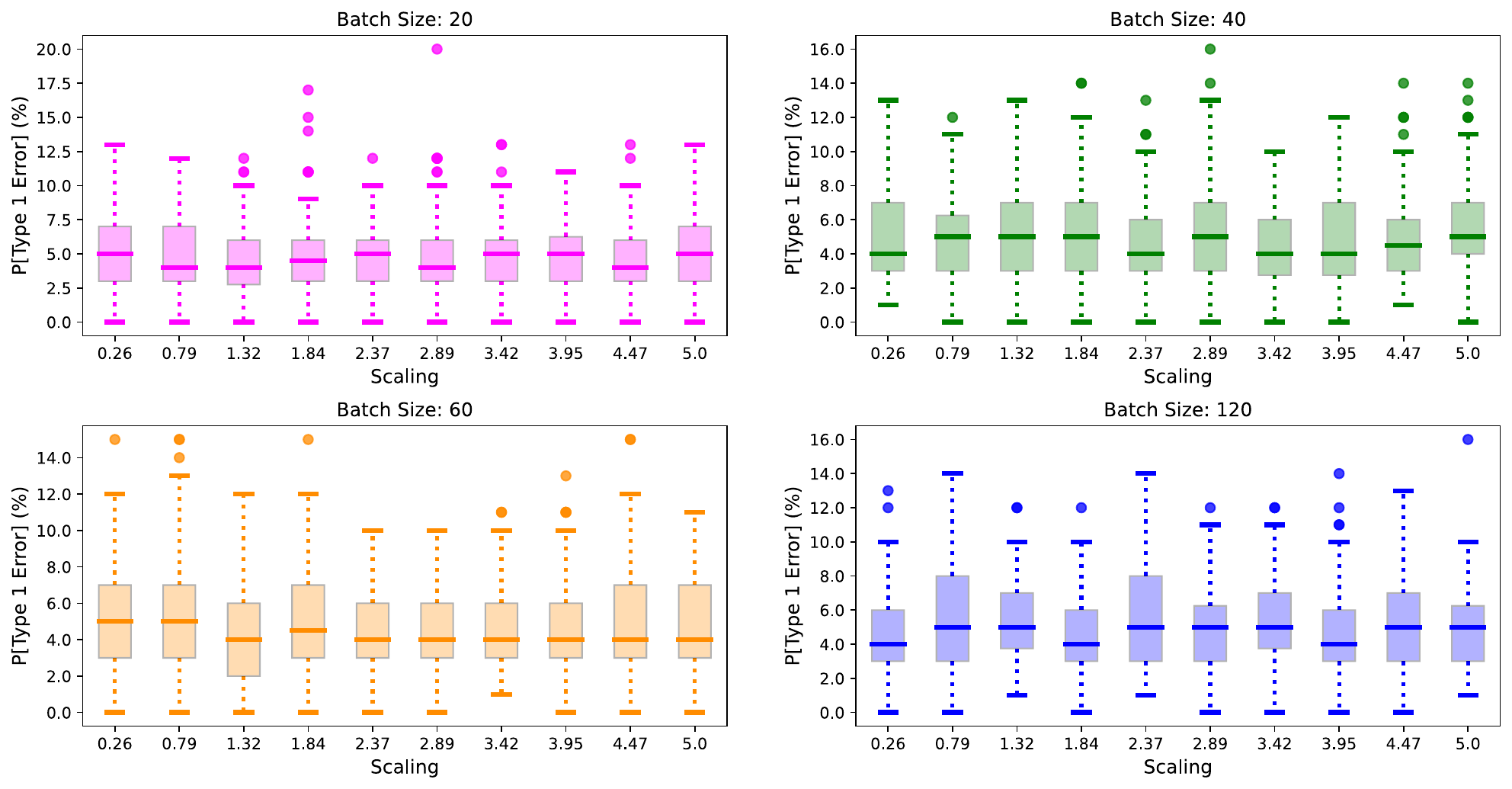}
    \caption{Unbiased}
    \label{fig:scaling_bm_list_type1_scalings_unbiased}
    \end{subfigure}

    \caption{Probability of a Type 1 error occurring as a function of sample size and scaling factor between two scaled Brownian motions.}
    \label{fig:scaling_bm_list_type1_scalings}
\end{figure}

These results demonstrate the importance of calibrating the signature kernel to focus on specific moments of the distributions. Apart from focusing on Type 2 errors, we ensured that the probability of a Type 1 error occurring did not increase as the scaling is altered. We did this by plotting the distributions of the probability of a Type 1 error occurring across various scaling factors (Fig.\ \ref{fig:scaling_bm_list_type1_scalings}). These plots show that this probability is independent of batch size and it averages at around $5\%$ in every scenario, corresponding to the significance level set for the test. These plots confirm that scaling does not negatively effect the probability of a Type 1 error occurring. This is consistent across both the biased and the unbiased estimator. \newline

We now study the effect of path scalings on higher dimensional paths. Consider the processes $\mathbf{X} = \left(\mathbf{X}_{t} = \left(t, S_{t}^{1}, \cdots, \right. \right.$ $\left. \left.S_{t}^{d-1}\right) \right)_{t}$ and $\mathbf{Y} = \left(\mathbf{Y}_{t} = \left(t, G_{t}^{1}, \cdots, G_{t}^{d-1}\right) \right)_{t}$. Each process pair $\left(\mathbf{S}^{j}, \mathbf{G}^{j}\right)$ is a scaled Brownian motion with parameters $\left(\sigma_{j}, \beta_{j}\right)$. We set $d = 5$ and used the following parameter values
\begin{itemize}
    \item $\left(\sigma_{1}, \beta_{1}\right) = \left(0.2, 0.3 \right)$;
    \item $\left(\sigma_{2}, \beta_{2}\right) = \left(0.5, 0.6 \right)$;
    \item $\left(\sigma_{3}, \beta_{3}\right) = \left( 0.7, 0.8 \right)$; and 
    \item $\left(\sigma_{4}, \beta_{4} \right) = \left( 0.1, 0.2 \right)$.
\end{itemize}
\noindent
Table \ref{tab:prob_type_2_dim_scaled_brownian} contains the probability of a Type 2 error occurring for various batch sizes and dimensions. A scaling factor of $2$ was applied to compute the probabilities. Scaling the paths does reduce the probability of a Type 2 error. In addition, scaled paths always had a lower probability than their original (unscaled) counterpart. More importantly, as the batch size increases, the probability does decrease when no scaling is applied since the sig-MMD is a metric and converges as sample size increases. The main effect of scaling the paths in this case is capturing distributional differences with smaller batch sizes. This is crucial when working in low-data regimes and/or working with limited computational resources.

\begin{table}[h]
  \caption{Likelihood of a Type 2 error occurring as dimension and batch size increase.}
  \label{tab:prob_type_2_dim_scaled_brownian}
  \centering
  \begin{tabular}{c@{\hskip 0.15cm}c@{\hskip 0.15cm}c@{\hskip 0.15cm}c@{\hskip 0.15cm}c@{\hskip 0.15cm}c@{\hskip 0.15cm}}
    \toprule
    Dimension & \thead{Sample \\ Size} & \thead{Bias \\ Scaling (\(\%\))}  & \thead{Bias \\ No Scaling (\(\%\))} & \thead{Unbiased \\ Scaling (\(\%\))} & \thead{Unbiased \\ No Scaling (\(\%\))} \\
    \midrule
    \midrule
    \multirow{6}{2em}{3} &16&\(77.8\)&\(90.5\)&\(87.8\)&\(92.2\)
    \\
    &\cellcolor{gray!25}32&\cellcolor{gray!25}\(72.1\)&\cellcolor{gray!25}\(89.9\)&\cellcolor{gray!25}\(84.6\)&\cellcolor{gray!25}\(92.6\)
    \\
    &64&\(58.2\)&\(87.4\)&\(75.3\)&\(90.5\)\\
    &\cellcolor{gray!25}128&\cellcolor{gray!25}\(33.2\)&\cellcolor{gray!25}\(84.0\)&\cellcolor{gray!25}\(50.3\)&\cellcolor{gray!25}\(88.1\)
    \\
    &256&\(6.80\)&\(70.2\)&\(15.2\)&\(76.1\)
    \\
    \midrule
    \midrule
    \multirow{6}{2em}{4} &16&\(77.4\)&\(86.8\)&\(88.2\)&\(92.3\)\\
    &\cellcolor{gray!25}32&\cellcolor{gray!25}\(68.4\)&\cellcolor{gray!25}\(85.4\)&\cellcolor{gray!25}\(83.9\)&\cellcolor{gray!25}\(91.3\)
    \\
    &64&\(53.6\)&\(83.4\)&\(75.6\)&\(89.3\)
    \\
    &\cellcolor{gray!25}128&\cellcolor{gray!25}\(31.0\)&\cellcolor{gray!25}\(73.8\)&\cellcolor{gray!25}\(57.2\)&\cellcolor{gray!25}\(83.5\)
    \\
    &256&\(8.13\)&\(52.4\)&\(24.5\)&\(68.8\)
    \\
    \midrule
    \midrule
    \multirow{6}{2em}{5} &16&\(61.4\)&\(86.6\)&\(88.3\)&\(92.4\)
    \\
    &\cellcolor{gray!25}32&\cellcolor{gray!25}\(49.4\)&\cellcolor{gray!25}\(85.3\)&\cellcolor{gray!25}\(83.2\)&\cellcolor{gray!25}\(91.3\)
    \\
    &64&\(28.9\)&\(82.0\)&\(72.8\)&\(89.1\)
    \\
    &\cellcolor{gray!25}128&\cellcolor{gray!25}\(6.9\)&\cellcolor{gray!25}\(73.5\)&\cellcolor{gray!25}\(42.6\)&\cellcolor{gray!25}\(84.5\)
    \\
    &256&\(0.1\)&\(49.3\)&\(4.4\)&\(67.1\)
  \end{tabular}
\end{table}

\subsection{Autoregressive Time Series Models} \label{sec:garch_example}

This example focuses on autoregressive conditional heteroskedasticity (ARCH) processes. Let $\mathbf{r} = \left(\mathbf{r}_{t}\right)$ be a stochastic process with conditional volatility modelled as a general autoregressive conditional heteroskedastic (GARCH) process \cite{garch_model_defn, garch_vol}, a class of ARCH models \cite{arch_model_defn}. The process $\mathbf{r}$ is described by the equations
\begin{align*}
    &\mathbf{r}_{t} = \mu + \epsilon_{t}  && \\\nonumber
    &\sigma_{t}^{2} = \omega + \alpha_{1} \epsilon_{t-1}^{2} + \beta_{1} \sigma_{t-1}^{2} && \\\nonumber
    &\epsilon_{t} = \sigma_{t}z_{t} && \\\nonumber
    &z_{t} \sim \mathcal{N} \left(0, 1\right).
\end{align*}

\begin{figure}[H]
     \centering
     \begin{subfigure}{0.45\textwidth}
         \centering
         \includegraphics[width=0.95\linewidth, height=0.2\textheight]{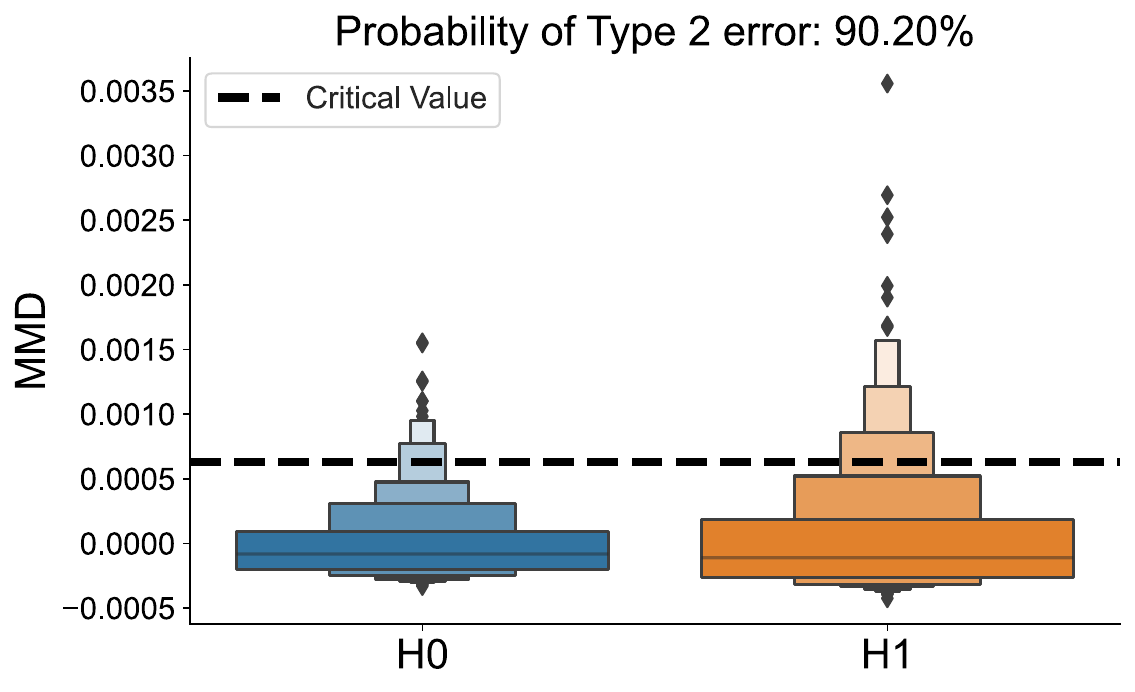}
         \caption{Without scaling.}
         \label{fig:garch_no_scaling}
     \end{subfigure}
     \begin{subfigure}{0.45\textwidth}
         \centering
         \includegraphics[width=0.95\linewidth, height=0.2\textheight]{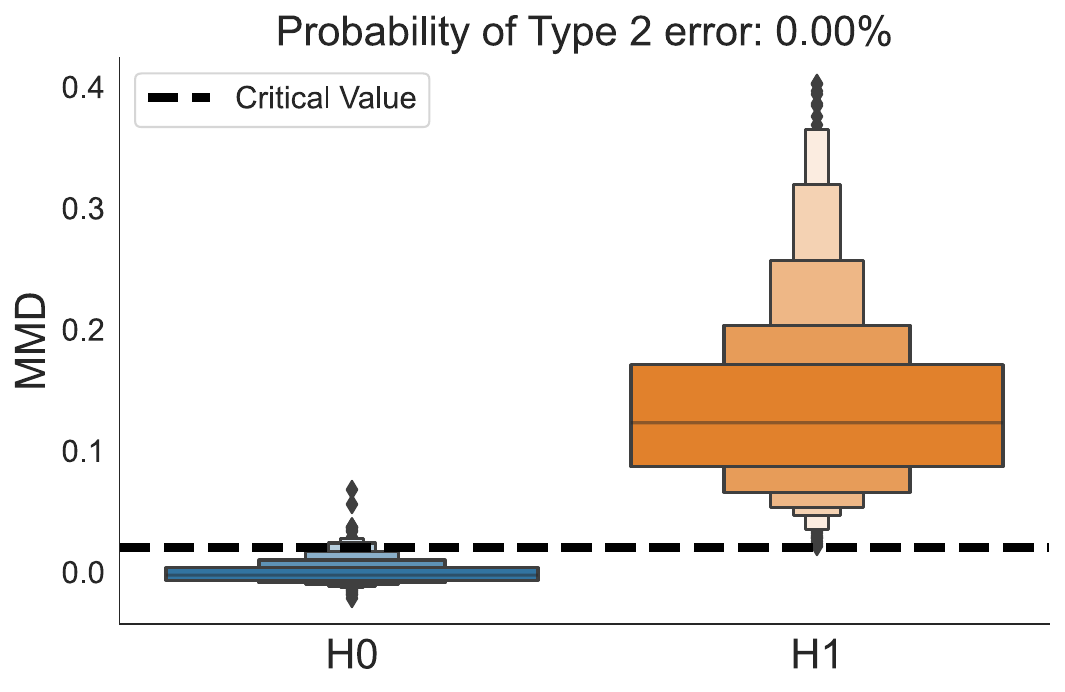}
         \caption{Scaling of $5.5$ was applied.}
         \label{fig:garch_scaling}
     \end{subfigure}
 	\caption{Null and alternative distributions of the $\phi$-MMD between two GARCH models. Batch size of $128$ was used and $500$ independent simulations were run.}
        \label{fig:garch_test}
\end{figure}

Two processes $\mathbf{r}_{1}, \mathbf{r}_{2}$ were simulated using the parameters in Table \ref{tab:garch_params}. As was done in the previous example (Section \ref{sec:brownian_motion_example}), the probability of a Type 2 error occurring in the two-sample hypothesis test was calculated through re-sampling techniques. Using the unbiased estimator, the computation resulted in a probability of $90.2\%$ (Fig.\ \ref{fig:garch_no_scaling}). Once again, path scaling was sufficient to reduce this probability. When a scaling of $5.5$ was applied, the probability of a Type 2 error occurring dropped to $0.0\%$ (Fig.\ \ref{fig:garch_scaling}). \newline

\begin{table}[h!]
\caption{GARCH model parameters.}
  \label{tab:garch_params}
  \centering
  \begin{tabular}{p{2cm}|p{2cm}p{2cm}p{2cm}p{2cm}}
\hline\noalign{\smallskip}
    Process & $\mu \left(\times 10^{-3}\right)$ & $\omega \left(\times 10^{-3} \right)$ & $\alpha_{1} \left( \times 10^{-2} \right)$  & $\beta_{1} \left(\times 10^{-2} \right)$\\
    \noalign{\smallskip}\hline \hline \noalign{\smallskip}
    $\mathbf{r}_{1}$ & $1.0$ & $3.8$ & $4$ & $4.2$  \\
    $\mathbf{r}_{2}$ & $5.0$ & $5.3$ & $8.0$  & $1.0$\\
    \noalign{\smallskip}\hline\noalign{\smallskip}
  \end{tabular}
\end{table}

\begin{figure}[H]
     \centering
     \begin{subfigure}{0.45\textwidth}
         \centering
         \includegraphics[width=0.95\linewidth, height=0.2\textheight]{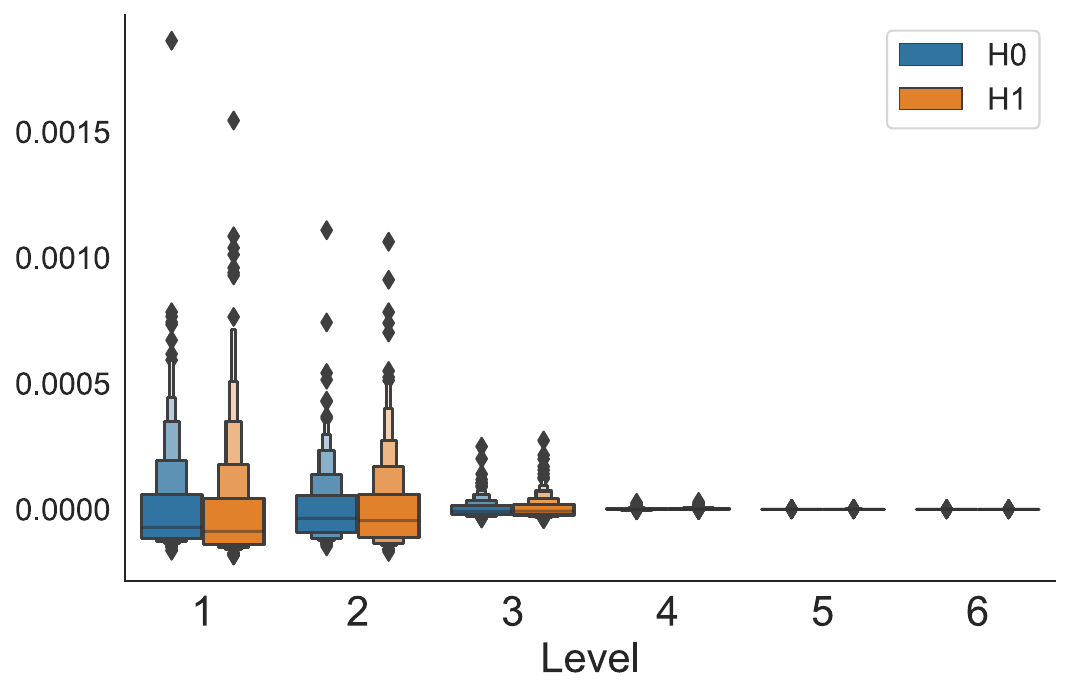}
         \caption{Without scaling.}
         \label{fig:garch_no_scaling_level_terms}
     \end{subfigure}
     \begin{subfigure}{0.45\textwidth}
         \centering
         \includegraphics[width=0.95\linewidth, height=0.2\textheight]{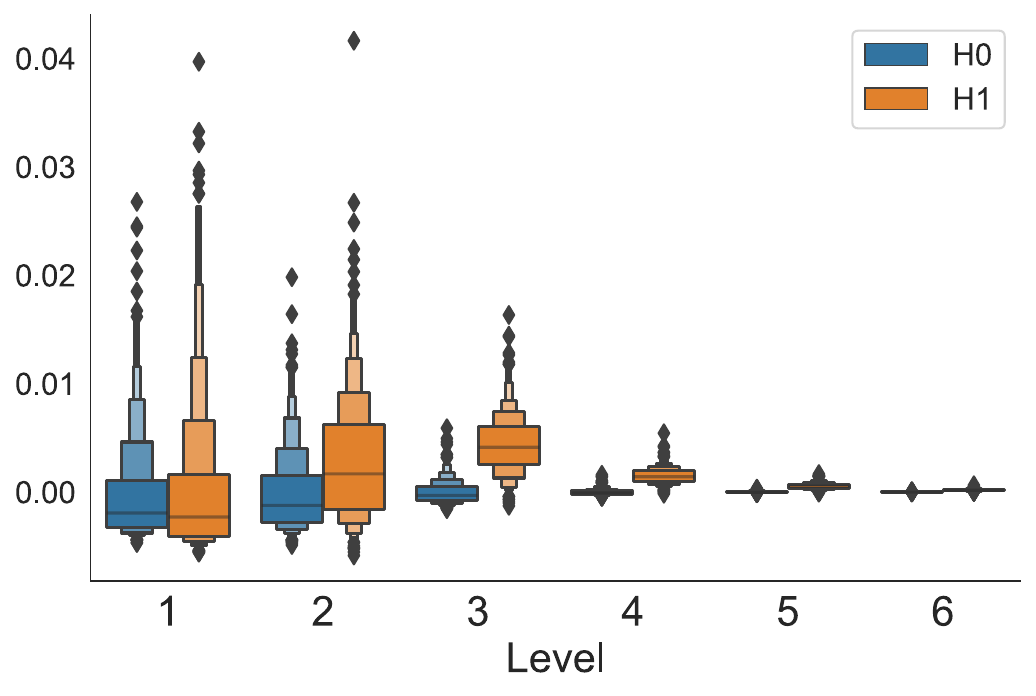}
         \caption{Scaling of $5.5$ was applied.}
         \label{fig:garch_scaling_level_terms}
     \end{subfigure}
 	\caption{Null and alternative distributions of the level contributions between two GARCH models. Batch size of $128$ was used and $500$ simulations were run.}
        \label{fig:garch_level_terms}
\end{figure}

Once again, the empirical distributions of the level contributions to the sig-MMD in the case of two GARCH models contained significant overlap (Fig.\ \ref{fig:garch_no_scaling_level_terms}). When a scaling of $5.5$ was applied, the distribution of level terms contained sufficient separation (Fig.\ \ref{fig:garch_scaling_level_terms}). Another important difference between the distributions of the level terms with and without scaling is that when scaling is used, the level contributions in the case of the alternative hypothesis have larger absolute value. As previously discussed, this contributed to reducing the probability of a Type 2 error occurring. \newline

\begin{figure}[H]
    \centering
    \begin{subfigure}{0.45 \textwidth}
        \includegraphics[width=1.0\linewidth, height=0.2\textheight]{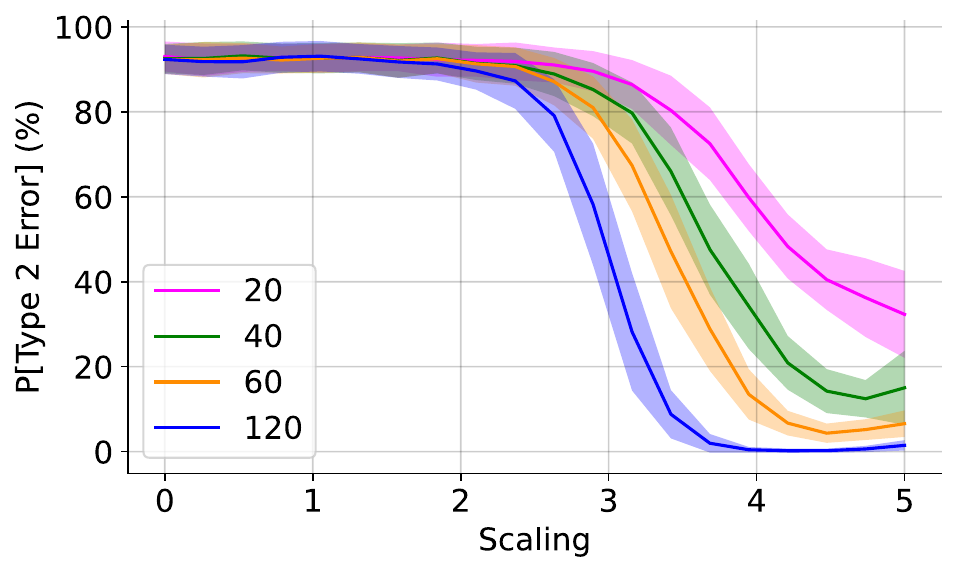}
        \caption{Biased}
        \label{fig:scaling_garch_list_type2_scalings_biased}
    \end{subfigure}
        \begin{subfigure}{0.45 \textwidth}
        \includegraphics[width=1.0\linewidth, height=0.2\textheight]{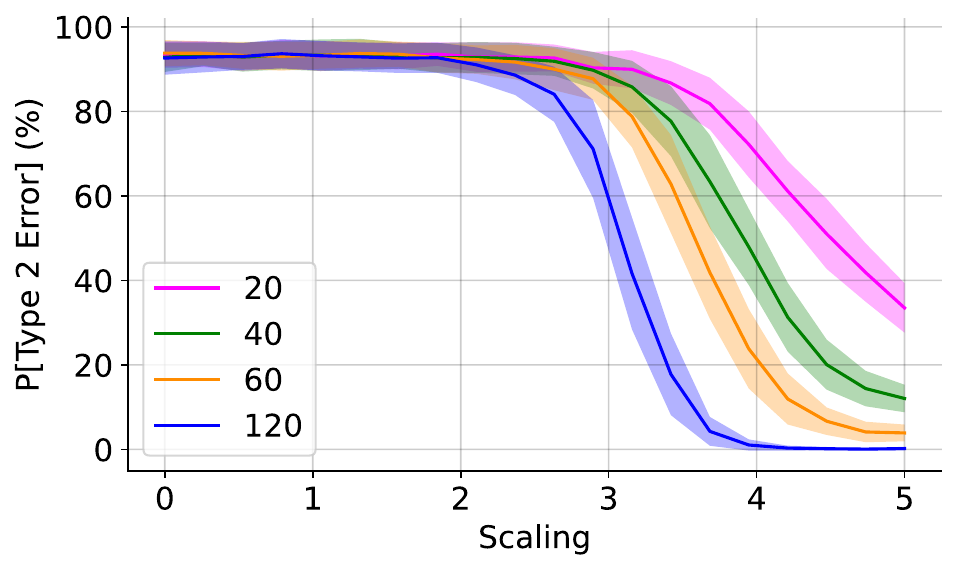}
        \caption{Unbiased}
        \label{fig:scaling_garch_list_type2_scalings_unbiased}
    \end{subfigure}
    \caption{Probability of a Type 2 error occurring between two GARCH models as a function of sample size.}
    \label{fig:scaling_garch_list_type2_scalings}
\end{figure}

\begin{figure}[h]
     \centering
     \begin{subfigure}{0.45\textwidth}
         \centering
         \includegraphics[width=0.95\linewidth, height=0.2\textheight]{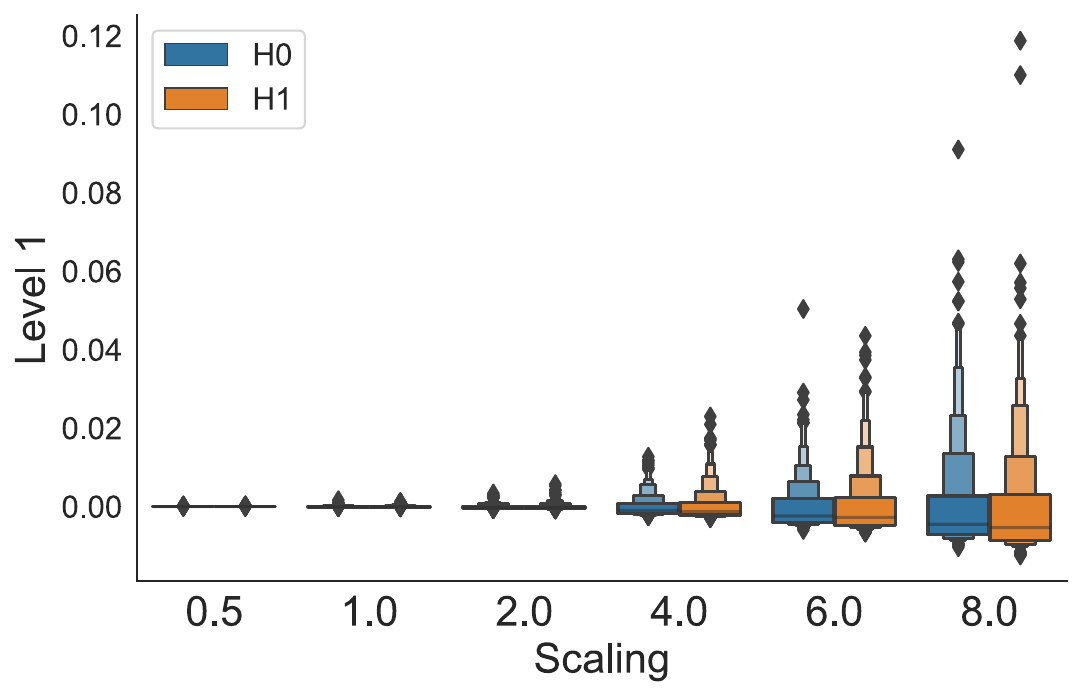}
         \caption{Level 1.}
         \label{fig:garch_level1}
     \end{subfigure}
     \begin{subfigure}{0.45\textwidth}
         \centering
         \includegraphics[width=0.95\linewidth, height=0.2\textheight]{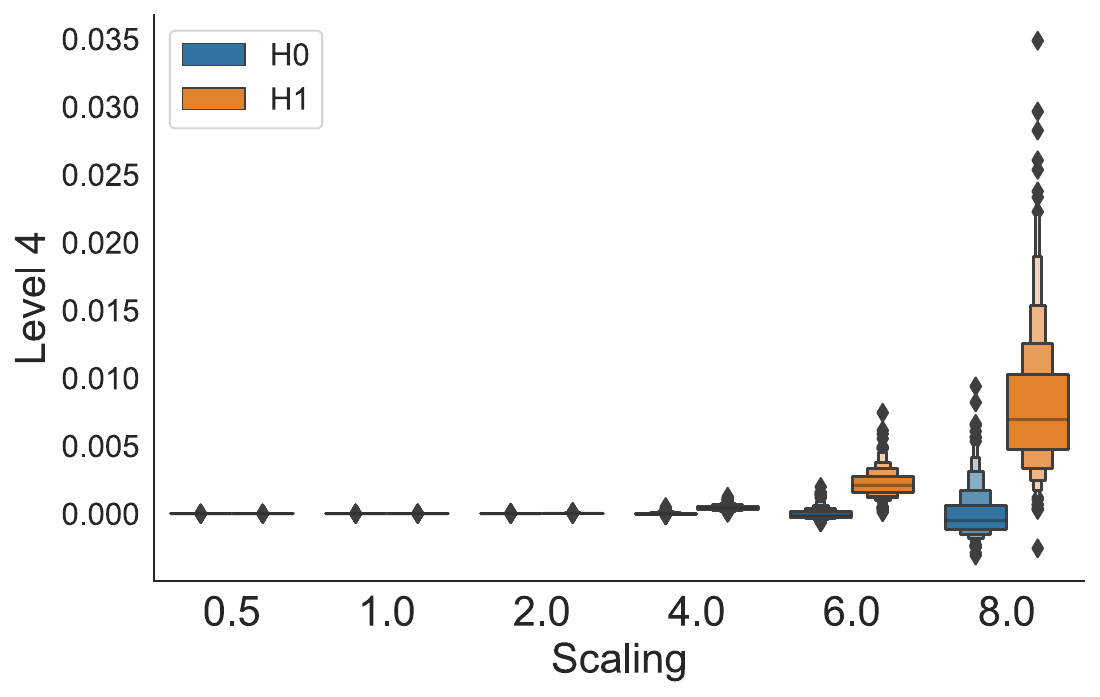}
         \caption{Level 4.}
         \label{fig:garch_level4}
     \end{subfigure}
    \begin{subfigure}{0.45\textwidth}
         \centering
         \includegraphics[width=0.95\linewidth, height=0.2\textheight]{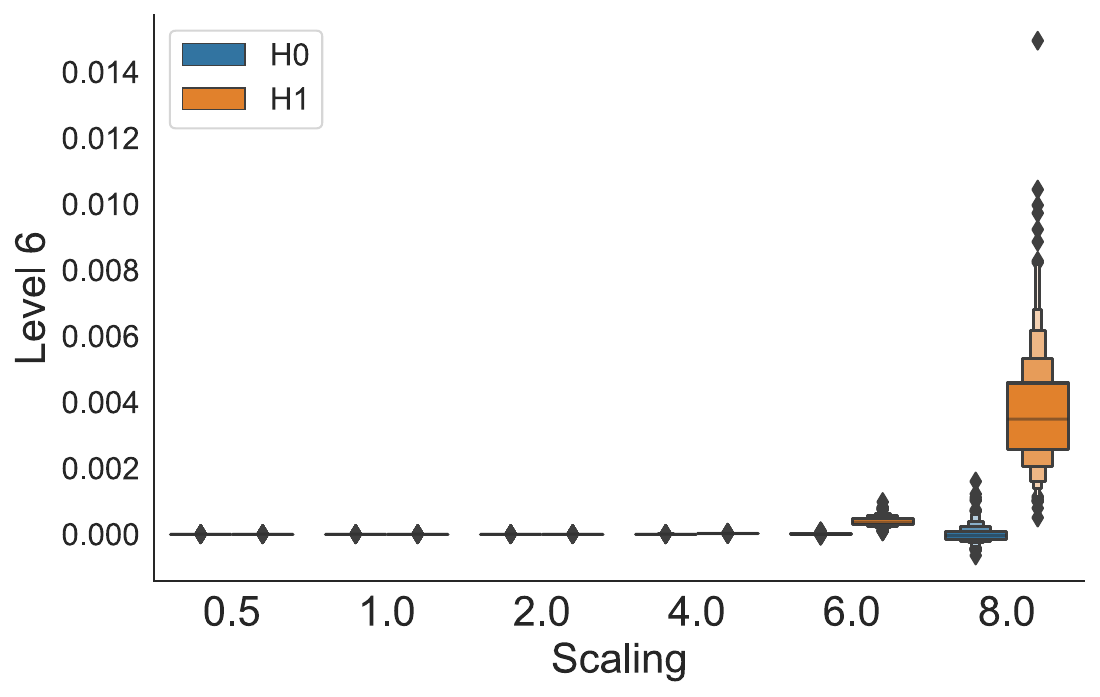}
         \caption{Level 6.}
         \label{fig:garch_level6}
     \end{subfigure}
 	\caption{Distribution of the level terms between two GARCH models as a function of scaling factor. Unbiased estimator was used.}
        \label{fig:garch_level_terms_scaling}
\end{figure}

All GARCH simulations were performed using a batch size of $128$. We once again plotted the probability of a Type 2 error occurring as the batch size increases and scaling factor increases. This is provided in Fig.\ \ref{fig:scaling_garch_list_type2_scalings}. Scaling does reduce the probability of a Type 2 error occurring. Comparing Fig.\ \ref{fig:scaling_garch_list_type2_scalings} with Fig.\ \ref{fig:scaling_bm_list}, we find that scaling has the common effect of reducing the probability of a Type 2 error occurring. However, a difference between the two is the scaling factor needed to reduce this probability. In the case of scaled Brownian motions, the probability started reducing when scalings slightly larger than $1$ were considered. In the case of GARCH models, a high probability was maintained up until a scaling of around $3.5$. The need for more aggressive scaling when working with GARCH models was confirmed in the distribution of the level terms depicted in Fig.\ \ref{fig:garch_level_terms_scaling}. Fig.\ \ref{fig:garch_level1} shows the first level contribution as a function of the scaling. Scaling does not alter the average contribution under either hypothesis. It mainly affects the variance of the level contribution. When considering the fourth and sixth level terms (Fig.\ \ref{fig:garch_level4} and Fig.\ \ref{fig:garch_level6} respectively), besides effect the variance the scaling also alters the mean contribution. However, this effect occurs at high scaling values. For example, in Fig.\ \ref{fig:garch_level4} there is a noticeable separation in distributions when considering scaling factors greater than $4.0$. \newline

\begin{figure}[h]
    \centering
    \includegraphics[width=1.0\textwidth, height=0.3\textheight]{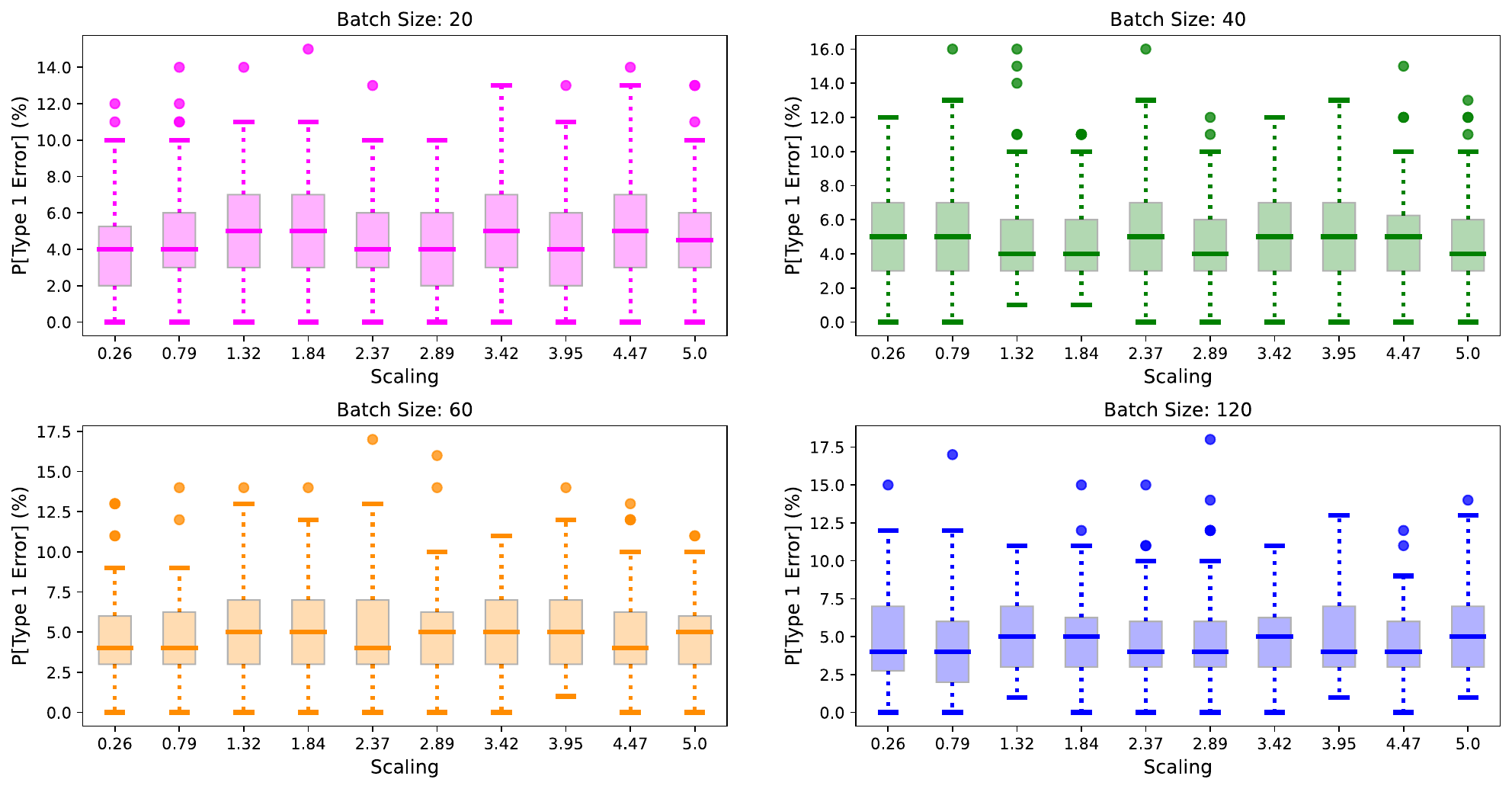}
    \caption{Probability of a Type 1 error occurring between two GARCH models as a function of sample size and scaling factor. Unbiased estimator was used.}
    \label{fig:scaling_garch_list_type1_scalings}
\end{figure}

We once again study the effect of scaling on the probability of a Type 1 error occurring. The distribution of the probability of a Type 1 error occurring using the $\phi$-MMD as a function of various scaling factors and batch sizes is plotted in Fig.\ \ref{fig:scaling_garch_list_type1_scalings}. On average, the probability remained stable at around the $5\%$ value. \newline

In this example, differences between processes were difficult to capture. This was confirmed by the level contributions plotted in Fig.\ \ref{fig:garch_level_terms} and Fig.\ \ref{fig:garch_level_terms_scaling} and the plot showing the effect of scaling on the probability of a Type 2 error occurring (Fig.\ \ref{fig:scaling_garch_list_type2_scalings}). Although the processes have different first two moments, higher-order terms of the signature were needed to capture differences.

\subsection{Mixture Models}

To understand the effect of scaling on the Type 2 error of the statistical test between two stochastic processes which differ in their third moment, we use a mixture of a geometric Brownian motion (GBM) \cite{black_scholes_paper} and an Ornstein-Uhlenbeck (OU) process \cite{ou_process_paper}. The GBM is described by the SDE
\begin{equation*}
    d\mathbf{S}_{t} = \mu \mathbf{S}_{t} dt + \sigma \mathbf{S}_{t} dW_{t}
\end{equation*}
\noindent
and the OU process is described by the SDE
\begin{equation*}
    d\mathbf{G}_{t} = -\theta \mathbf{G}_{t} dt + \tilde{\sigma} dB_{t}
\end{equation*}
\noindent
where $\mu, \theta, \sigma, \tilde{\sigma} > 0$ and $\{W_{t}\}, \{B_{t}\}$ are Brownian motions. Consider the process
\begin{equation*}
    d\mathbf{X}_{t} = d\mathbf{S}_{t} + d\mathbf{G}_{t}
\end{equation*}
and assume $\{W_{t}\}, \{B_{t}\}$ are uncorrelated Brownian motions. By choosing specific parameter values, the marginal distributions of $X_{T}, Y_{T}$ with $T=1$ satisfy
\begin{equation*}
    \mathbb{E} \left[X_{T} \right] \approx \mathbb{E} \left[Y_{T}\right],~~\mathbb{E} \left[X_{T}^{2}\right] \approx \mathbb{E} \left[Y_{T}^{2}\right],~~\text{and}~~
    \mathbb{E} \left[X_{T}^{3}\right] \neq \mathbb{E} \left[Y_{T}^{3}\right].
\end{equation*}
   
\begin{figure}[H]
     \centering
     \begin{subfigure}{0.45\textwidth}
         \centering
         \includegraphics[width=0.95\linewidth, height=0.2\textheight]{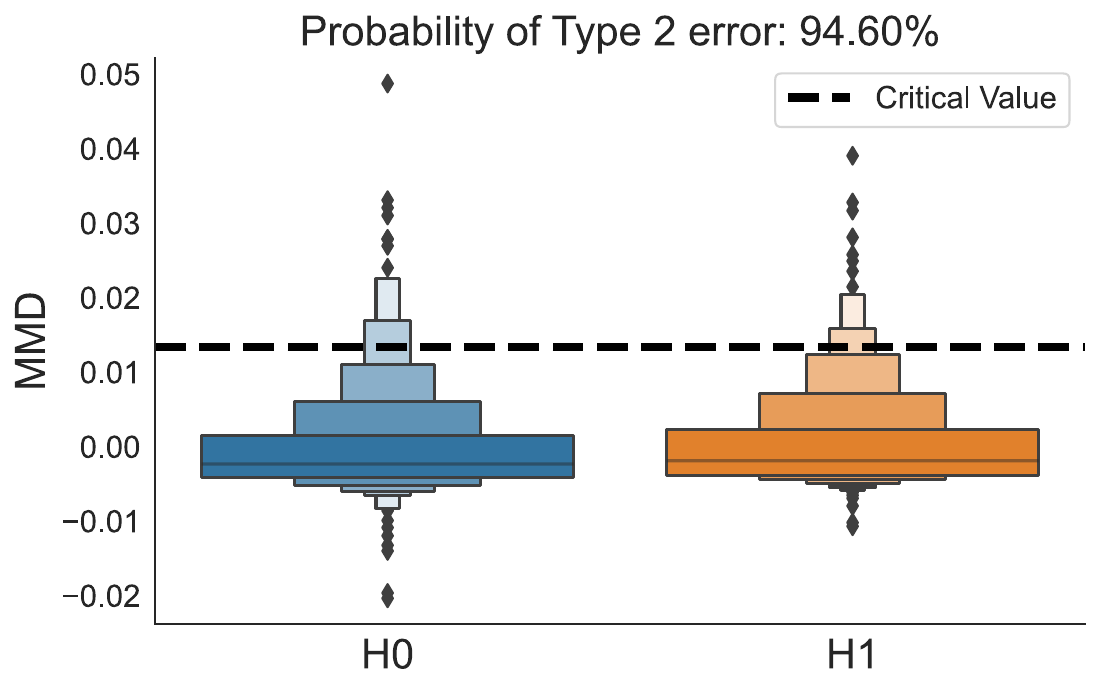}
         \caption{No scaling.}
         \label{fig:test_simulation1_noscaling}
     \end{subfigure}
     \begin{subfigure}{0.45\textwidth}
         \centering
         \includegraphics[width=0.95\linewidth, height=0.2\textheight]{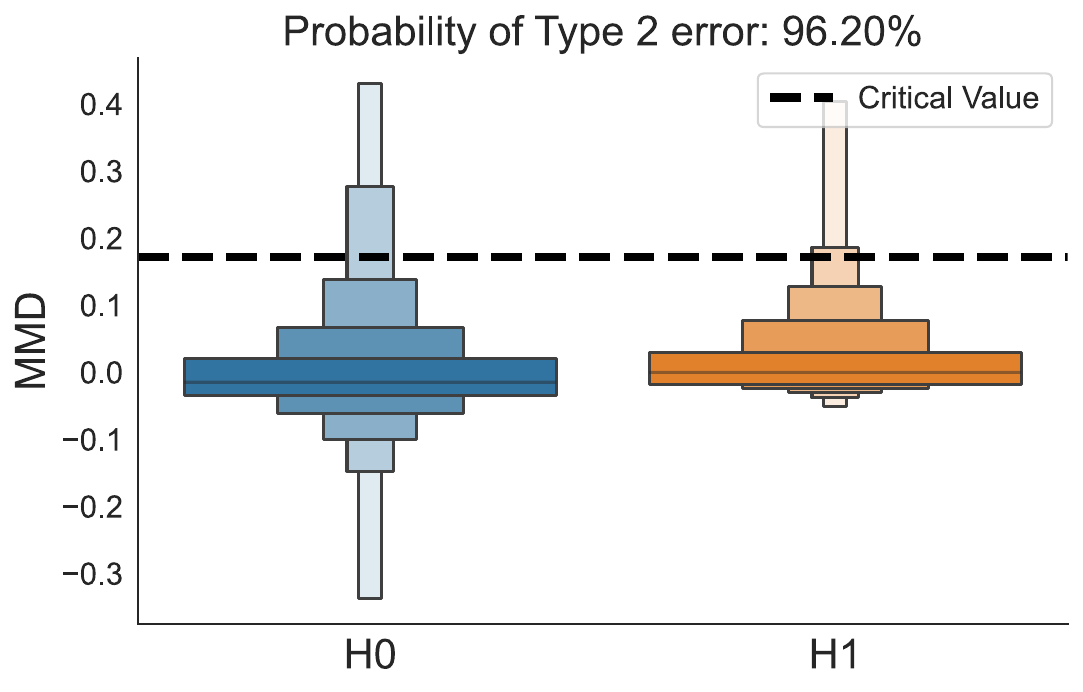}
         \caption{Scaling of $2.0$ was applied.}
         \label{fig:test_simulation1_scaling2}
     \end{subfigure}
    \caption{Null and alternative distributions of the $\phi$-MMD between two mixture models. Batch size of $128$ and the unbiased estimator were used.}
    \label{fig:third_moment_dist_sim1}
\end{figure}

\begin{figure}[h]
     \centering
     \begin{subfigure}{0.45\textwidth}
         \centering
         \includegraphics[width=0.95\linewidth, height=0.2\textheight]{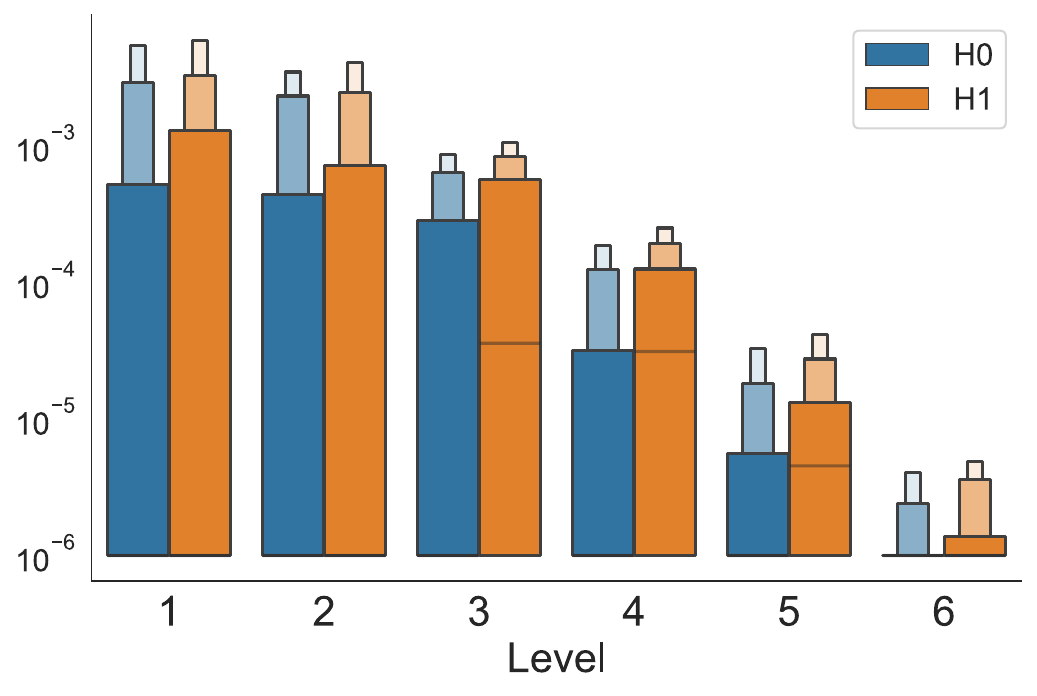}
         \caption{Batch size of $128$.}
         \label{fig:mixture_level_128}
     \end{subfigure}
     \begin{subfigure}{0.45\textwidth}
         \centering
         \includegraphics[width=0.95\linewidth, height=0.2\textheight]{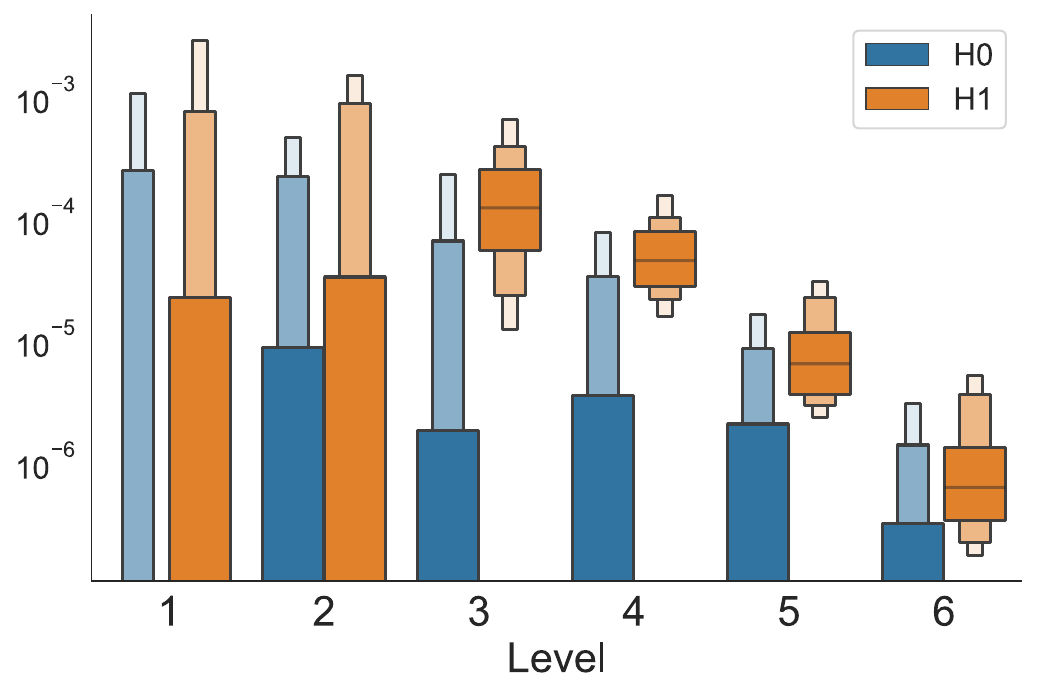}
         \caption{Batch size of $512$.}
         \label{fig:mixture_level_512}
     \end{subfigure}
    \begin{subfigure}{0.45\textwidth}
         \centering
         \includegraphics[width=0.95\linewidth, height=0.2\textheight]{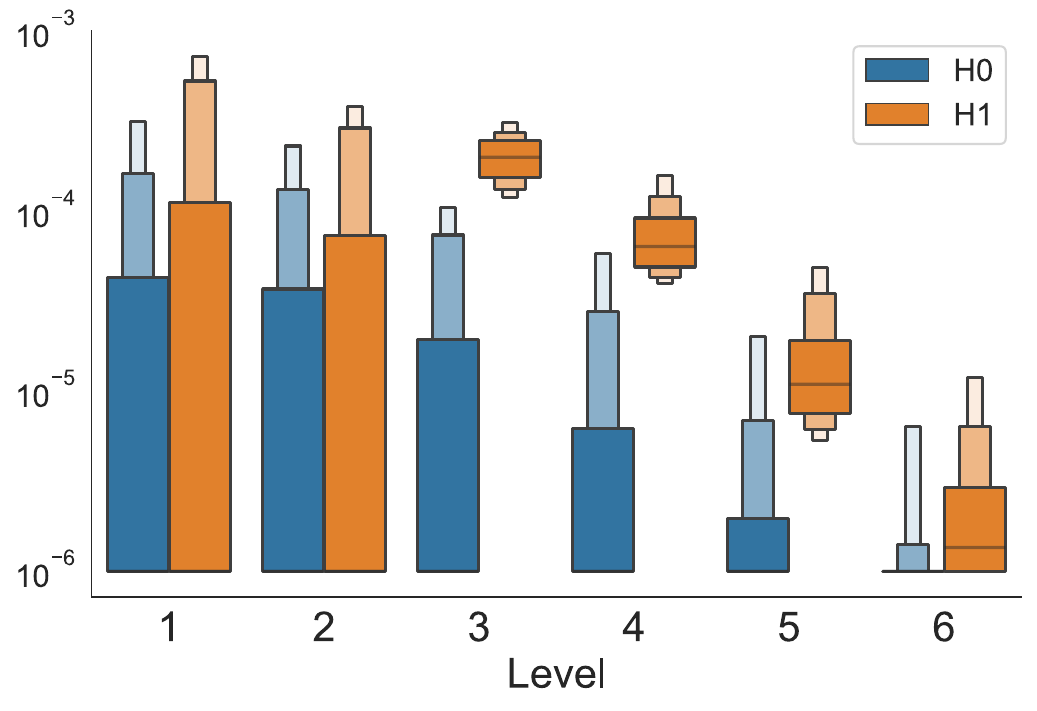}
         \caption{Batch size of $2{,}000$.}
         \label{fig:mixture_level_2000}
     \end{subfigure}
 	\caption{Null and alternative distributions of the level contributions between two mixture models as a function of batch size. $100$ independent simulations were run. The level contributions are plotted on a logarithmic scale.}
        \label{fig:mixture_levels_no_scaling}
\end{figure}

\begin{figure}[h]
     \centering
     \begin{subfigure}{0.45\textwidth}
         \centering
         \includegraphics[width=0.95\linewidth, height=0.2\textheight]{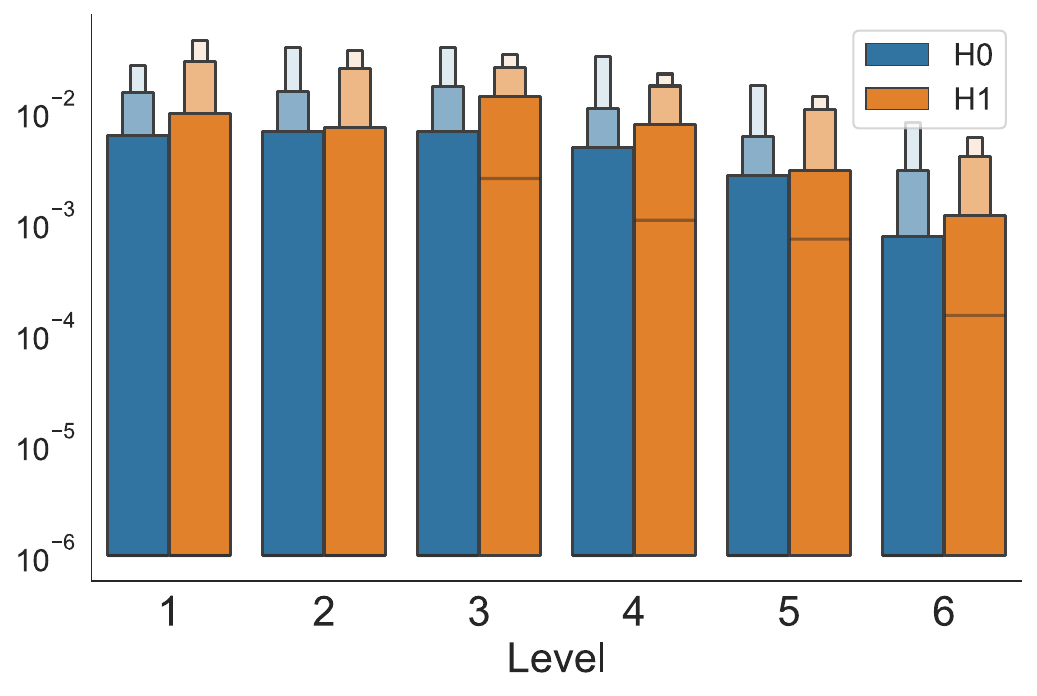}
         \caption{Batch size of $128$.}
         \label{fig:mixture_level_128_scaling}
     \end{subfigure}
     \begin{subfigure}{0.45\textwidth}
         \centering
         \includegraphics[width=0.95\linewidth, height=0.2\textheight]{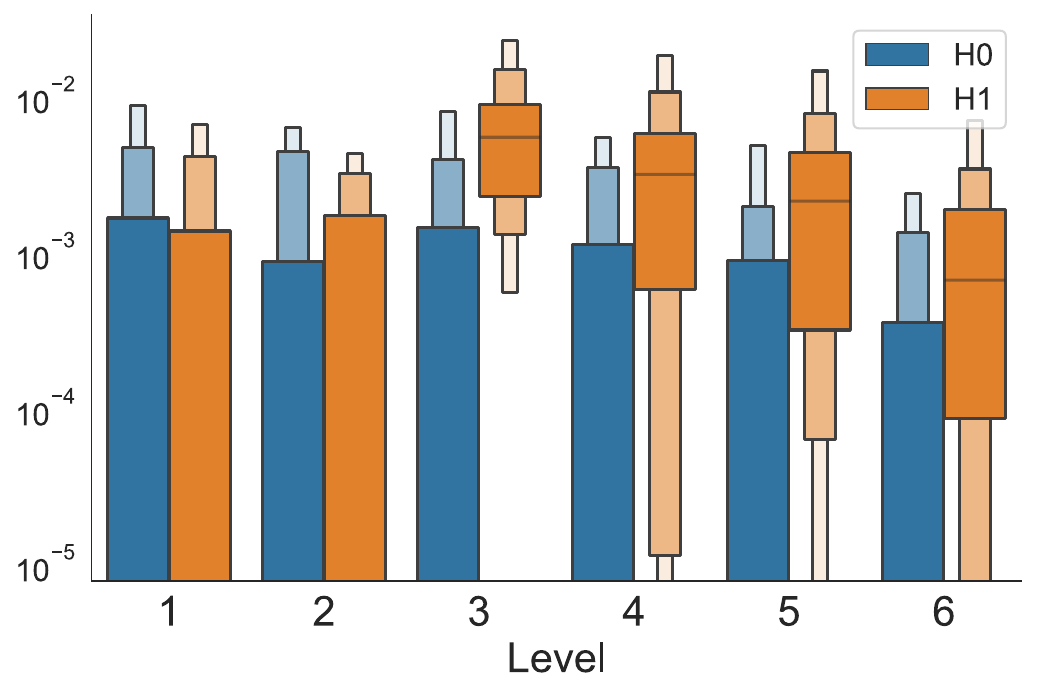}
         \caption{Batch size of $512$.}
         \label{fig:mixture_level_512_scaling}
     \end{subfigure}
    \begin{subfigure}{0.45\textwidth}
         \centering
         \includegraphics[width=0.95\linewidth, height=0.2\textheight]{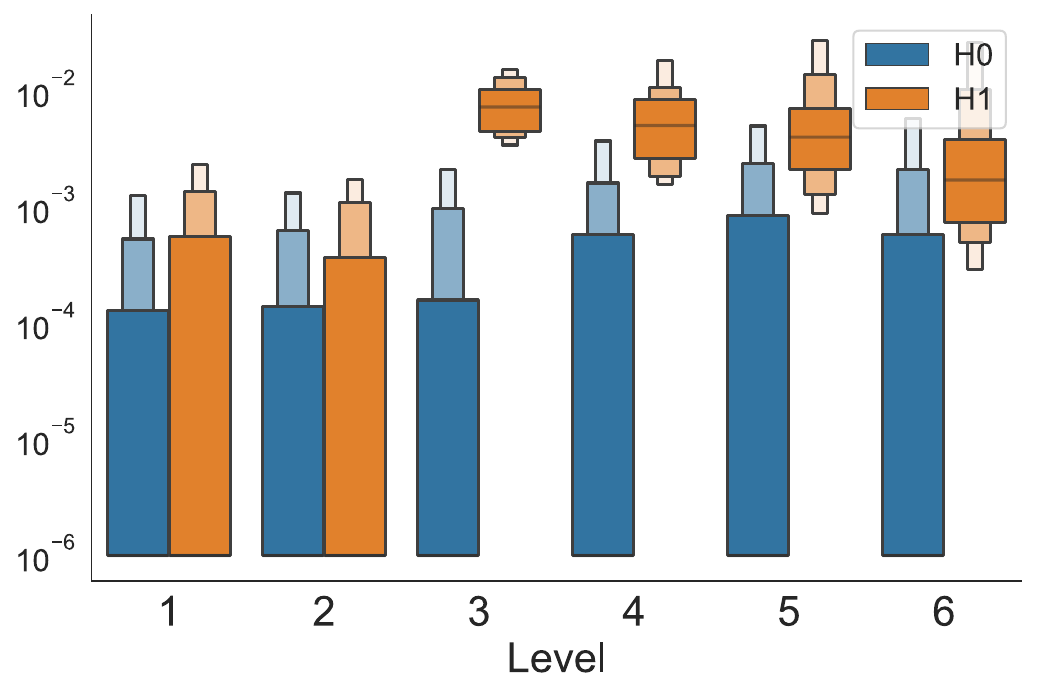}
         \caption{Batch size of $2{,}000$.}
         \label{fig:mixture_level_2000_scaling}
     \end{subfigure}
 	\caption{Null and alternative distributions of the level contributions between two mixture models as a function of batch size. A scaling of $2$ was applied. $100$ independent simulations were run. The values are plotted on a logarithmic scale.}
        \label{fig:mixture_levels_scaling}
\end{figure}

We simulate processes according to the parameters specified in Table \ref{tab:third_moment_params}. We use paths of length $30$ and a batch size of $128$. The probability of a Type 2 error occurring was $94.6\%$. This corresponds to the overlap in distributions in Fig.\ \ref{fig:test_simulation1_noscaling}. Scaling the paths did not reduce this probability (Fig.\ \ref{fig:test_simulation1_scaling2}). To understand the reason for this, we plot the level contributions. In Fig.\ \ref{fig:mixture_level_128}, at each level, there is overlap between distributions. Increasing the batch size to $512$ and $2{,}000$, the distributions diverge at levels $3$, $4$, and $5$. Therefore, the sig-MMD can distinguish between these processes, but it cannot distinguish when using a few paths. Re-plotting the distribution of the level contributions with a scaling of $2$ applied (Fig.\ \ref{fig:mixture_levels_scaling}), we still note significant overlap between the null and alternative distributions at a batch size of $128$. By increasing the batch size we have less overlap at levels $3$, $4$, $5$, and $6$. As a result of scaling, the absolute value of the level contributions under the alternative hypothesis do not decrease as rapidly compared to Fig.\ \ref{fig:mixture_levels_no_scaling}. \newline

\begin{table}[H]
  \caption{Mixture model parameters used to perform simulations. }
  \label{tab:third_moment_params}
  \centering
  \begin{tabular}{c|cccccc}
    \toprule
    Process & $\mu$ & $\theta$ & $\sigma$  & $\tilde{\sigma}$ & $G_{0}$ & $S_{0}$\\
    \midrule
    \midrule
    $\mathbf{X}$ & $0.3$ & $0.3$ & $0.5$ & $0.5$ & $0.75$ & $1.0$  \\
    $\mathbf{Y}$ &  $0.3$ & $0.3$ & $0.3$  & $0.84$ & $0.75$  & $1.0$   \\
    \bottomrule
  \end{tabular}
\end{table}

To overcome these challenges, we use three additional techniques; namely \newline

\noindent
\textbf{Path Standardisation:} We standardise the paths using the technique adopted in \cite{issa2023non}. If $\mathbf{X}_{t}$ is a path, we consider the standardised path
\begin{equation*}
    \widehat{\mathbf{X}}_{t} \coloneqq \left(\mathbf{X}_{t} - \mu_{T}\right) / \sigma_{T}
\end{equation*}
\noindent
where $\mu_{T}, \sigma_{T}$ are the mean and standard deviation of $\mathbf{X}$ at the terminal time. \newline

\noindent
\textbf{Lead-lag transformation:} Let $0 = t_{0} < t_{1/2} < \cdots < t_{l-1/2} < t_{l} = T$ be a partition of $\left[0, T\right]$. The lead-lag transformation \cite{sig_mmd_example, lead_lag_1, sig_ml_3} of a $1$-dimensional path $\mathbf{X}_{t}$ is the $2$-dimensional path $\left(\mathbf{X}^{\text{Lead}}_{t}, \mathbf{X}^{\text{Lag}}_{t}\right)$ defined by linear interpolation on the points
\begin{equation*}
    \mathbf{X}_{t_{i/2}}^{\text{Lead}} \coloneqq \begin{cases}
        \mathbf{X}_{t_{j}}~~& \text{if }i=2j \\
        \mathbf{X}_{t_{j+1}}~~& \text{if } i=2j+1
    \end{cases},~~~~~
    \mathbf{X}_{t_{i/2}}^{\text{Lag}} \coloneqq \begin{cases}
        \mathbf{X}_{t_{j}}~~& \text{if }i=2j \\
        \mathbf{X}_{t_{j}}~~& \text{if } i=2j+1
    \end{cases}.
\end{equation*}

\noindent
\textbf{Feature space transformation:} To gain more expressive power, the paths are lifted to the infinite dimensional Hilbert space $H^{\prime} \left(\sigma_{\text{RBF}}^{2}\right)$ associated with the radial basis function (RBF) kernel with smoothing parameter $\sigma_{\text{RBF}} > 0$. For $\mathbf{x}, \mathbf{y} \in \mathbb{R}^{d}$, the RBF kernel is defined by
\begin{equation*}
    k \left(\mathbf{x}, \mathbf{y}; \sigma_{\text{RBF}} \right) \coloneqq \exp \left(- \frac{\left| \left| \mathbf{x} - \mathbf{y} \right| \right|_{2}^{2}}{\sigma_{\text{RBF}}^{2}} \right)
\end{equation*}
\noindent
where $\left| \left| \cdot \right| \right|_{2}$ denotes the Euclidean norm. Using the Riesz representation theorem, there exists a mapping $\zeta_{\sigma_{\text{RBF}}} \colon \mathbb{R}^{d} \to H^{\prime}$ such that
\begin{equation*}
    k \left(\mathbf{x}, \mathbf{y}; \sigma_{\text{RBF}} \right) = \langle \zeta_{\sigma_{\text{RBF}}} \left(\mathbf{x}\right), \zeta_{\sigma_{\text{RBF}}} \left(\mathbf{y}\right) \rangle_{H^{\prime}}. 
\end{equation*}
The path $\mathbf{X}_{t}$ is lifted to the infinite dimensional path $k_{\mathbf{X}} \colon t \mapsto \zeta_{\sigma_{\text{RBF}}} \left(\mathbf{X}_{t}\right)$. Since the lifted path is infinite dimensional, it is not possible to work directly with this path. However, using the kernel trick, we work directly with the RBF kernel instead. Computing the signature kernel between two lifted paths can be done using either dynamic programming as described in \cite{kernel_sequential_data}, or directly through the PDE approach \cite{sig_kernel_pde_paper}. An alternative to directly computing the RBF kernel consists of approximating the RBF kernel using random Fourier features \cite{rffs, rff_sig}. An important consideration when working with lifted paths is that scaling the original paths does not necessarily translate to weighting the level contributions to the signature kernel since the signature kernel is now computed between the lifted paths. Therefore, we need to ensure that we are scaling the path $\zeta_{\sigma} \left(\mathbf{X}_{t}\right)$ and not the path $\mathbf{X}_{t}$. \newline

\begin{figure}[H]
     \centering
     \begin{subfigure}{0.45\textwidth}
         \centering
         \includegraphics[width=0.95\linewidth, height=0.2\textheight]{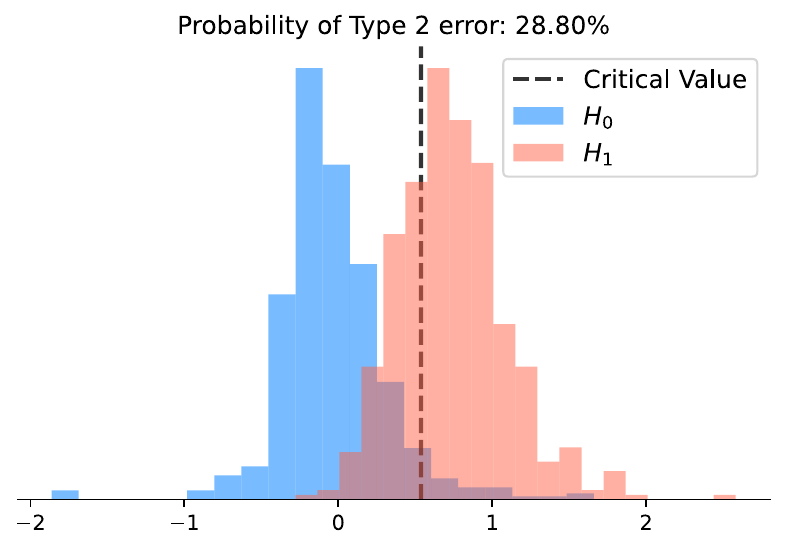}
         \caption{No scaling.}
         \label{fig:test_simulation1_rbf_noscaling}
     \end{subfigure}
     \begin{subfigure}{0.45\textwidth}
         \centering
         \includegraphics[width=0.95\linewidth, height=0.2\textheight]{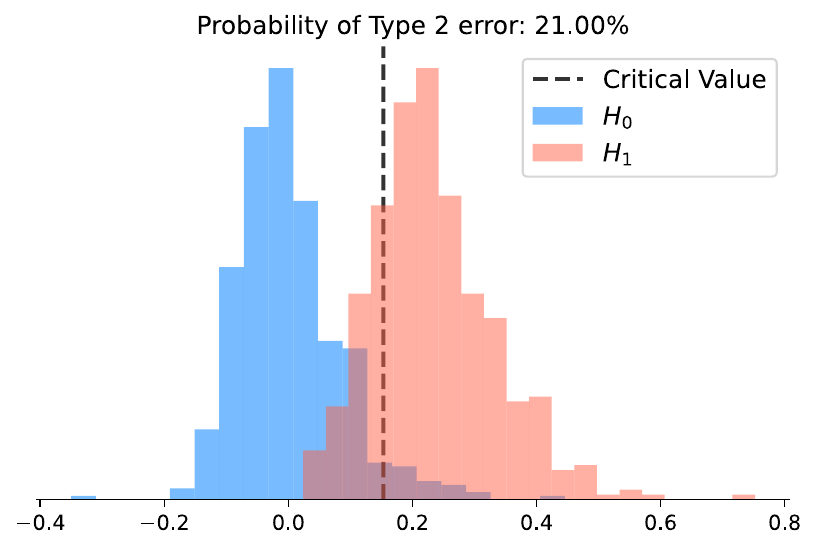}
         \caption{Scaling of $0.8$ was applied.}
         \label{fig:test_simulation1_rbf_scaling2}
     \end{subfigure}
    \caption{Null and alternative distributions of the $\phi$-MMD between two mixture models. Path standardisation and lead-lag transformations were performed. The RBF smoothing parameter was set to $\sigma_{\text{RBF}}=\sqrt{0.5}$. Batch size of $128$ was used.}
    \label{fig:third_moment_dist_sim1_rbf}
\end{figure}

Keeping the batch size fixed at $128$, the paths were first standardised using the above procedure. Performing the hypothesis test with lead-lag paths lifted to the space $H^{\prime} \left(0.5\right)$, the probability of a Type 2 error occurring reduced from $94.6\%$ to $28.8\%$ (Fig.\ \ref{fig:test_simulation1_rbf_noscaling}). By applying scaling, this was further reduced to $21.0\%$. \newline

\begin{figure}[H]
    \centering
    \begin{subfigure}{0.45 \textwidth}
        \includegraphics[width=1.0\linewidth, height=0.2\textheight]{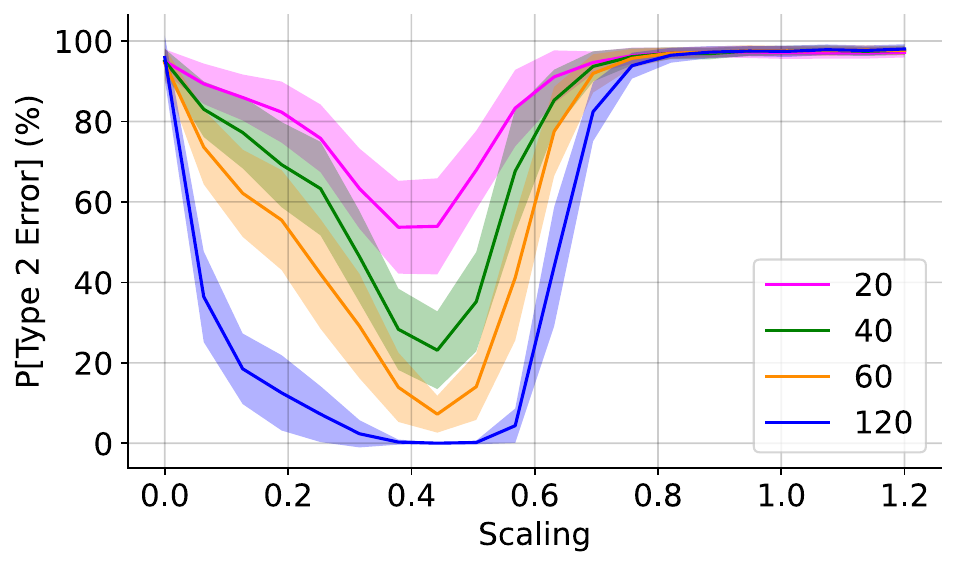}
        \caption{Biased - $\sigma_{\text{RBF}}=\sqrt{0.1}$}
        \label{fig:type2_scenario1_rbf_scaling_biased_sigma01}
    \end{subfigure}
    \begin{subfigure}{0.45 \textwidth}
        \includegraphics[width=1.0\linewidth, height=0.2\textheight]{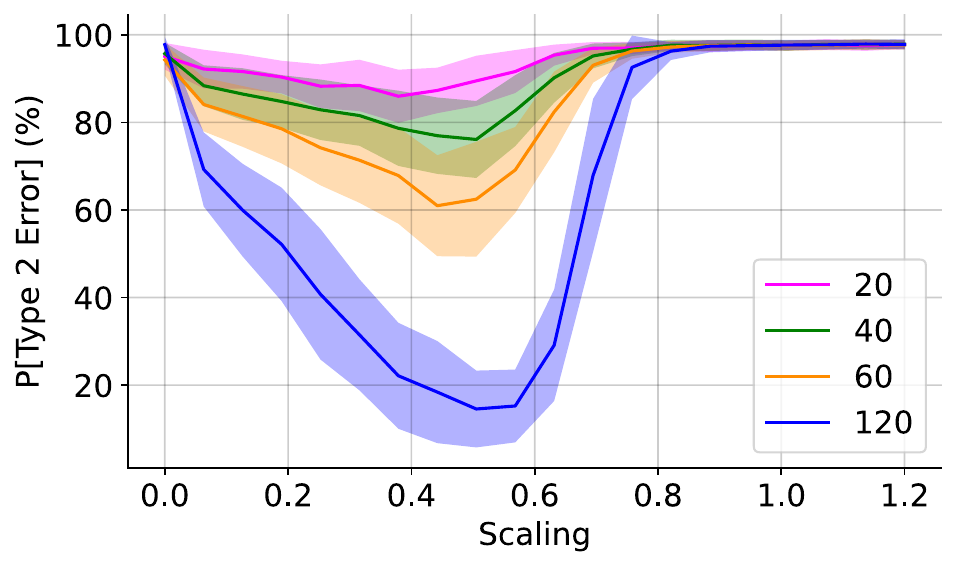}
        \caption{Biased - $\sigma_{\text{RBF}}=\sqrt{0.5}$}
        \label{fig:type2_scenario1_rbf_scaling_biased_sigma05}
    \end{subfigure}
    \begin{subfigure}{0.45 \textwidth}
        \includegraphics[width=1.0\linewidth, height=0.2\textheight]{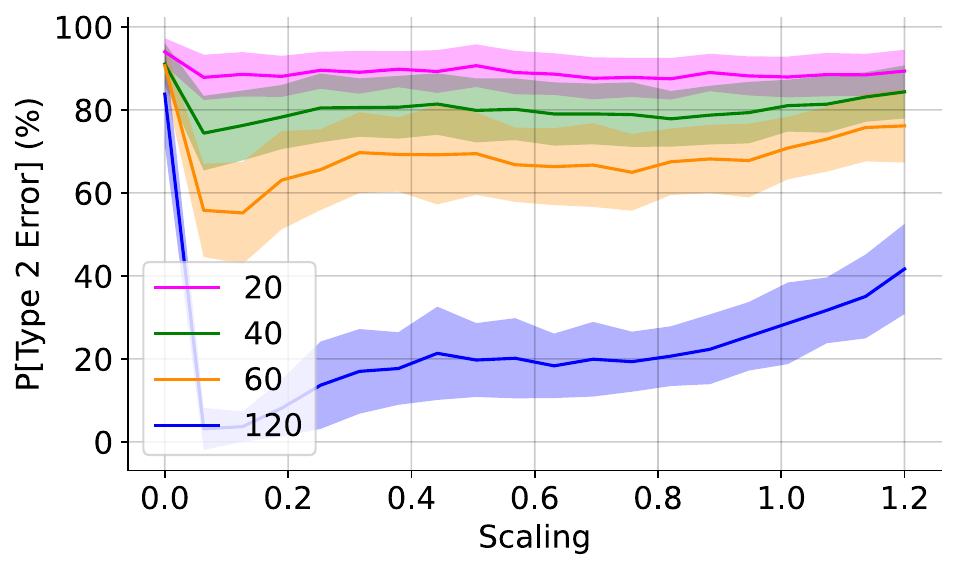}
        \caption{Unbiased - $\sigma_{\text{RBF}}=\sqrt{0.1}$}
        \label{fig:type2_scenario1_rbf_scaling_unbiased_sigma01}
    \end{subfigure}
    \begin{subfigure}{0.45 \textwidth}
        \includegraphics[width=1.0\linewidth, height=0.2\textheight]{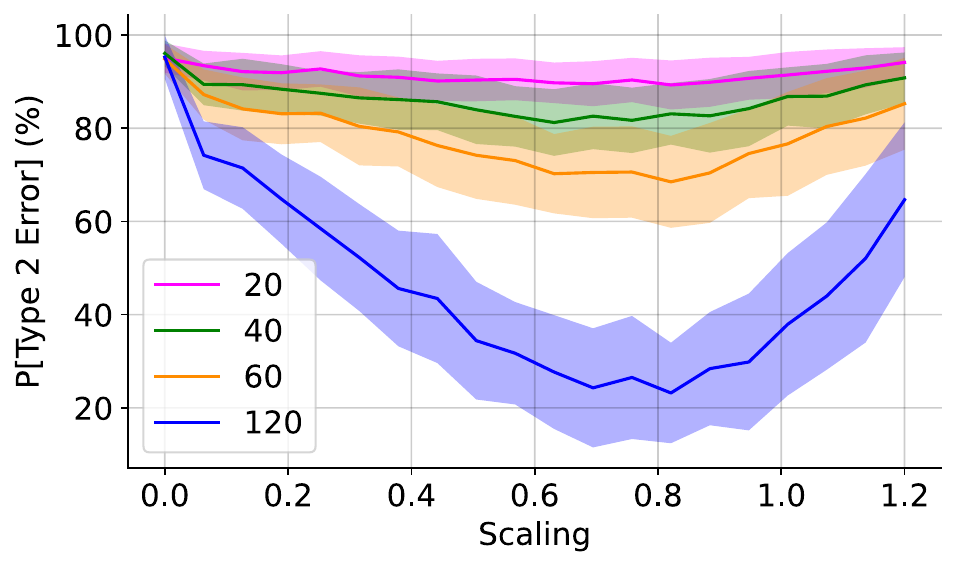}
        \caption{Unbiased - $\sigma_{\text{RBF}}=\sqrt{0.5}$}
        \label{fig:type2_scenario1_rbf_scaling_unbiased_sigma05}
    \end{subfigure}
    \caption{Probability of a Type 2 error occurring between two mixture models as a function of sample size.}
    \label{fig:type2_scenario1_rbf_scaling}
\end{figure}

\begin{figure}[h]
    \centering
    \includegraphics[width=1.0\textwidth, height=0.3\textheight]{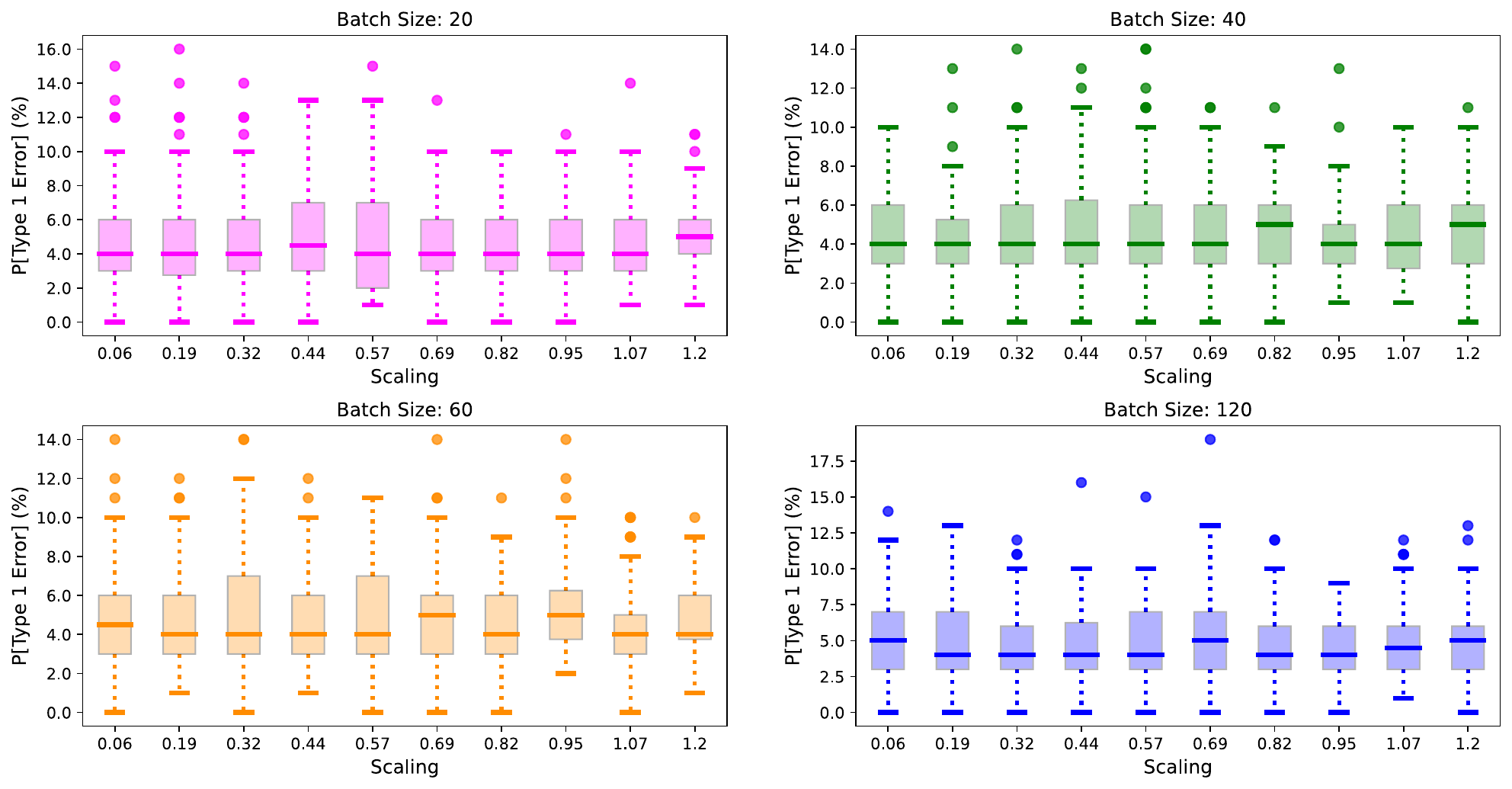}
    \caption{Probability of a Type 1 error occurring between two mixture models as a function of sample size and scaling factor. Biased estimator with $\sigma_{\text{RBF}}=\sqrt{0.5}$.}
    \label{fig:scaling_mixturemodel_scenario1_list_type1_scalings}
\end{figure}

To understand the relationship between path scalings and Type 2 error applied to lifted paths, we plot the probability of a Type 2 error as a function of the scaling factor and the batch size (Fig.\ \ref{fig:type2_scenario1_rbf_scaling}). Whenever the probability was significantly reduced, the curve of probabilities had a parabolic shape. A difference between the plots in Fig.\ \ref{fig:type2_scenario1_rbf_scaling} and the plots in Fig.\ \ref{fig:scaling_bm_list} and Fig.\ \ref{fig:scaling_garch_list_type2_scalings} is that, in Fig.\ \ref{fig:scaling_bm_list} and Fig.\ \ref{fig:scaling_garch_list_type2_scalings} there was a scaling which reduced the probability of Type 2 error occurring. This was not necessarily the case when performing the two-sample hypothesis test between mixture models (see Fig.\ \ref{fig:type2_scenario1_rbf_scaling_unbiased_sigma05}). Moreover, we once again show that the probability of a Type 1 error occurring is unaffected by the scaling factor used (Fig.\ \ref{fig:scaling_mixturemodel_scenario1_list_type1_scalings}). 

\section{Uncontrolled Environment} \label{sec:real_world}

In Section \ref{sec:numerical_examples}, we had full control over the null and alternative hypotheses, i.e.\ we knew whether the null or alternative hypothesis held prior to performing the test. In practice, we are not in a controlled environment and need to determine which hypothesis holds. As demonstrated through the simulations, finding a robust test setup suitable in all cases is not possible. We presented various techniques for tailoring the two-sample test depending on the collections of sample paths. In practice, finding an appropriate setup is similar to hyperparameter optimisation in a ML context. As is the case in most ML tasks, task performance is highly dependent on specific configurations (learning rates, network architecture, etc.). We propose optimising the scaling factor in a similar way, mimicking the optimisation procedure for certain hyperparameters in a ML context. Since we are focused on two-sample testing, the hyperparameters are optimised for a low probability of a Type 2 error occurring. We demonstrated various test configurations and techniques\footnote{There could be other techniques. We presented the ones we find most useful in practice.} which prove to be effective in the context of two-sample testing using path signatures. Since we have shown that the probability of a Type 1 error occurring is largely unaffected across all techniques presented, by iterating over the various techniques, we are optimising the test for Type 2 errors whilst maintaining a low rate of Type 1 errors. When performing hyperparameter optimisation in the context of two-sample hypothesis testing, one would need to split the data into two parts. The first part is used to fine-tune the hyperparameters with respect to some criterion and the second part of the data is used to perform the hypothesis test \cite{first_order_mmd_distinguish, train_test_split_3, train_test_split_2, mmd_agg, train_test_split_1}.  \newline 

\begin{figure}[h]
     \centering
     \begin{subfigure}{0.45\textwidth}
         \centering
         \includegraphics[width=0.95\linewidth, height=0.2\textheight]{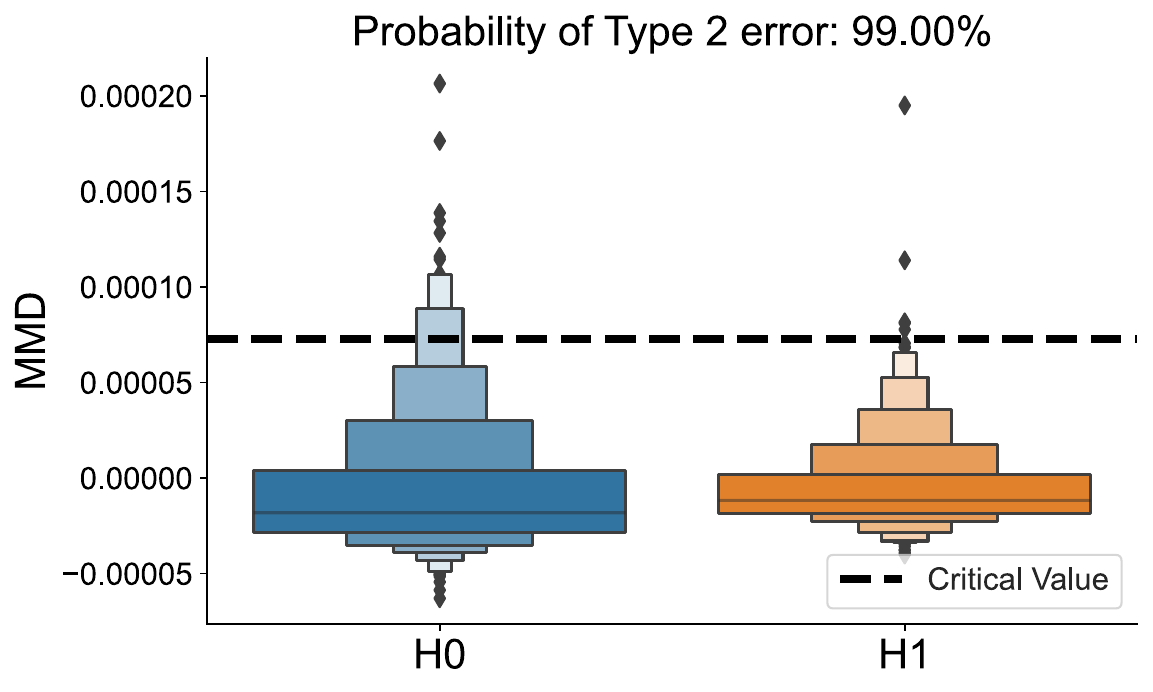}
         \caption{No scaling.}
         \label{fig:real_world_linear_noscaling}
     \end{subfigure}
     \begin{subfigure}{0.45\textwidth}
         \centering
         \includegraphics[width=0.95\linewidth, height=0.2\textheight]{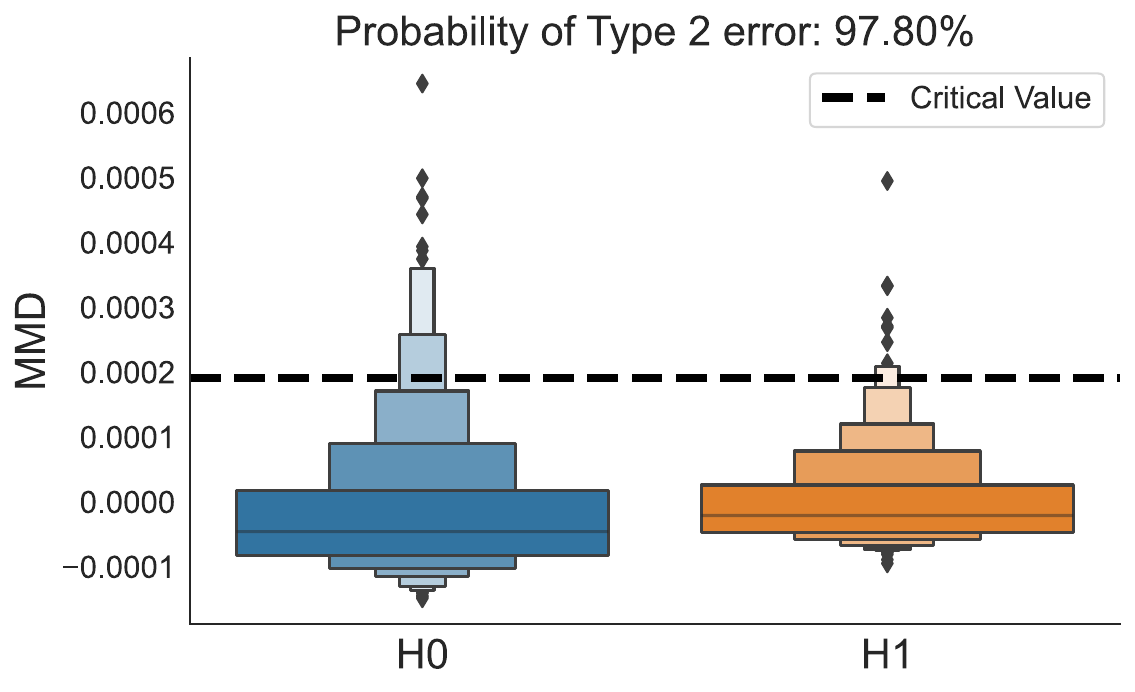}
         \caption{$\sigma_{\text{RBF}} = 1$.}
         \label{fig:real_world_rbf_noscaling}
     \end{subfigure}
    \begin{subfigure}{0.45\textwidth}
         \centering
         \includegraphics[width=0.95\linewidth, height=0.2\textheight]{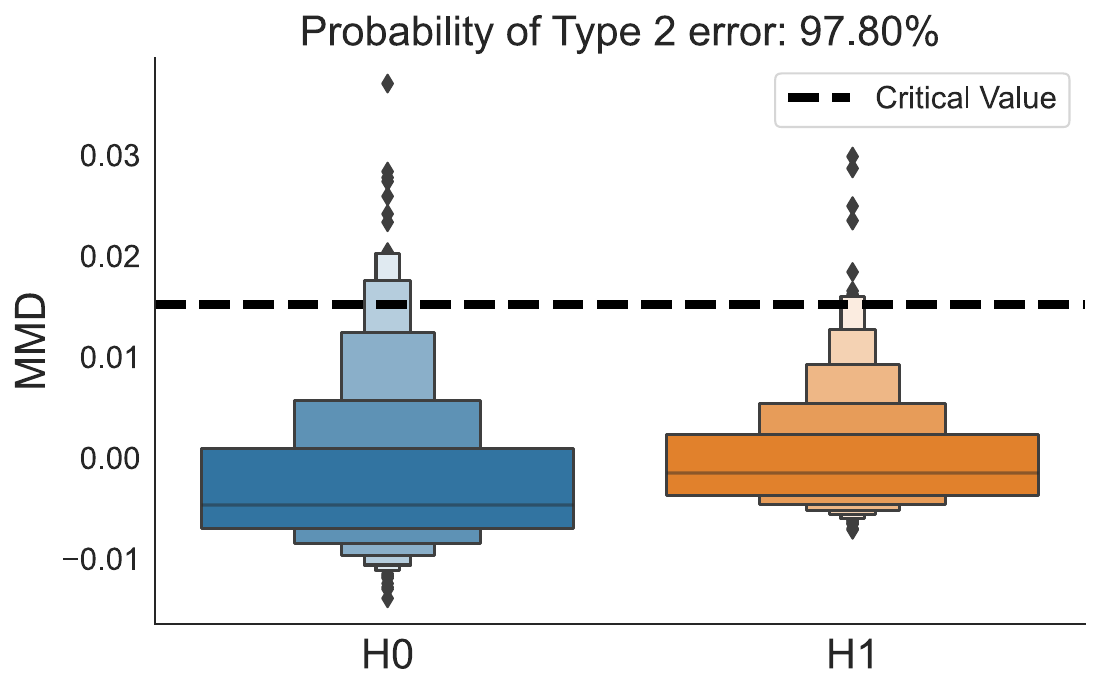}
         \caption{$\sigma_{\text{RBF}} = 1$, scaling$=5$.}
         \label{fig:real_world_rbf_scaling_5}
     \end{subfigure}
 	\caption{Null and alternative distributions of the $\phi$-MMD. $500$ independent simulations were run. Batch size of $128$ and unbiased estimator were used.}
        \label{fig:real_world_noleadlag}
\end{figure}

We now demonstrate how to apply these techniques in an uncontrolled environment. We would like to determine whether the time series of returns corresponding to baskets of assets have the same distribution. Each basket consists of assets from the same sector. The sectors we consider are the technology sector and consumer based sector. We use $5$ assets from each sector and for each asset we compute the percentage return. The return time series corresponding to each asset is split into multiple time series, each consisting of $15$ consecutive observations (approximately 3 weeks of returns)\footnote{The sub-sampled time series have no overlapping observations.}. Aggregating all length $15$ time series of all assets within the same basket (sector) constitutes the available samples from the stochastic process corresponding to that basket of assets. We use data from 10 September 2014 to 8 September 2024. This is an uncontrolled setting since we do not know in advance whether the returns have the same distribution. The data was split into two parts according to the ratio $80{:}20$. $80\%$ of the data was used to calibrate the hyperparameters of the test. We first start the calibration procedure by using the raw returns without any scaling applied. The probability of a Type 2 error occurring in this setup was calculated at $99.0\%$ (Fig.\ \ref{fig:real_world_linear_noscaling}). Performing a feature transformation using a RBF kernel with smoothing parameter set to $1$, we then performed the two-sample test again. This feature transformation had no effect on the probability of a Type 2 error occurring (Fig.\ \ref{fig:real_world_rbf_noscaling}). This was also the case when a scaling of $5$ was applied (Fig.\ \ref{fig:real_world_rbf_scaling_5}). \newline

\begin{figure}[h]
     \centering
     \begin{subfigure}{0.45\textwidth}
         \centering
         \includegraphics[width=0.95\linewidth, height=0.2\textheight]{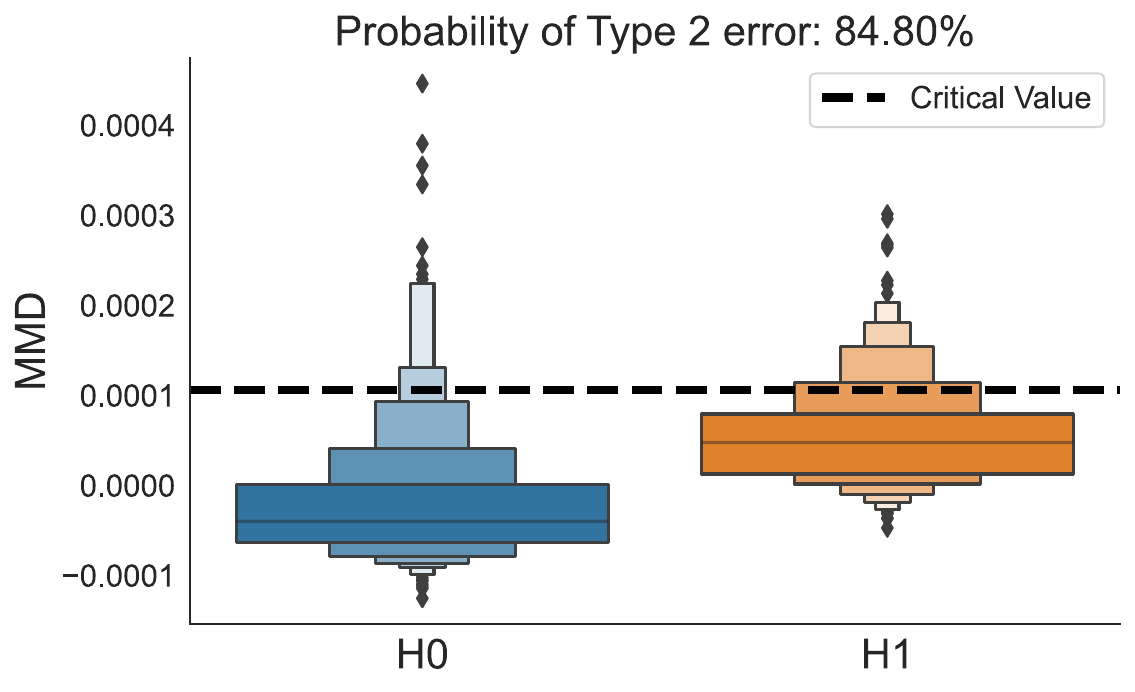}
         \caption{No scaling applied.}
         \label{fig:real_world_linear_noscaling_leadlag}
     \end{subfigure}
     \begin{subfigure}{0.45\textwidth}
         \centering
         \includegraphics[width=0.95\linewidth, height=0.2\textheight]{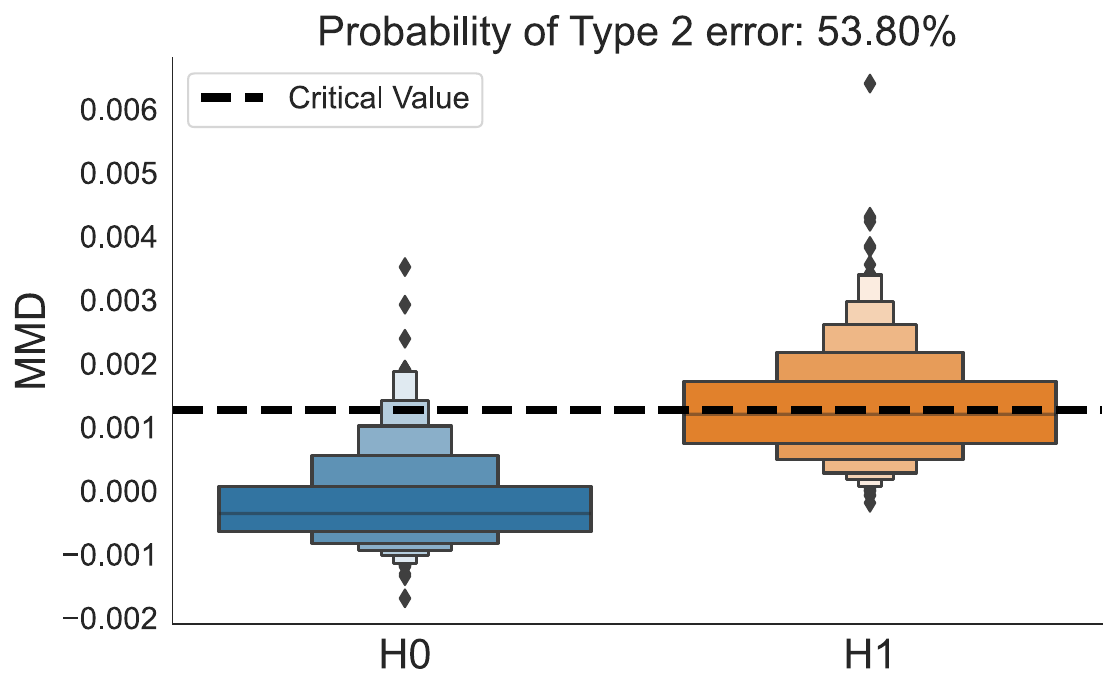}
         \caption{$\sigma_{\text{RBF}} = \sqrt{0.5}$.}
         \label{fig:real_world_rbf_noscaling_leadlag}
     \end{subfigure}
    \begin{subfigure}{0.45\textwidth}
         \centering
         \includegraphics[width=0.95\linewidth, height=0.2\textheight]{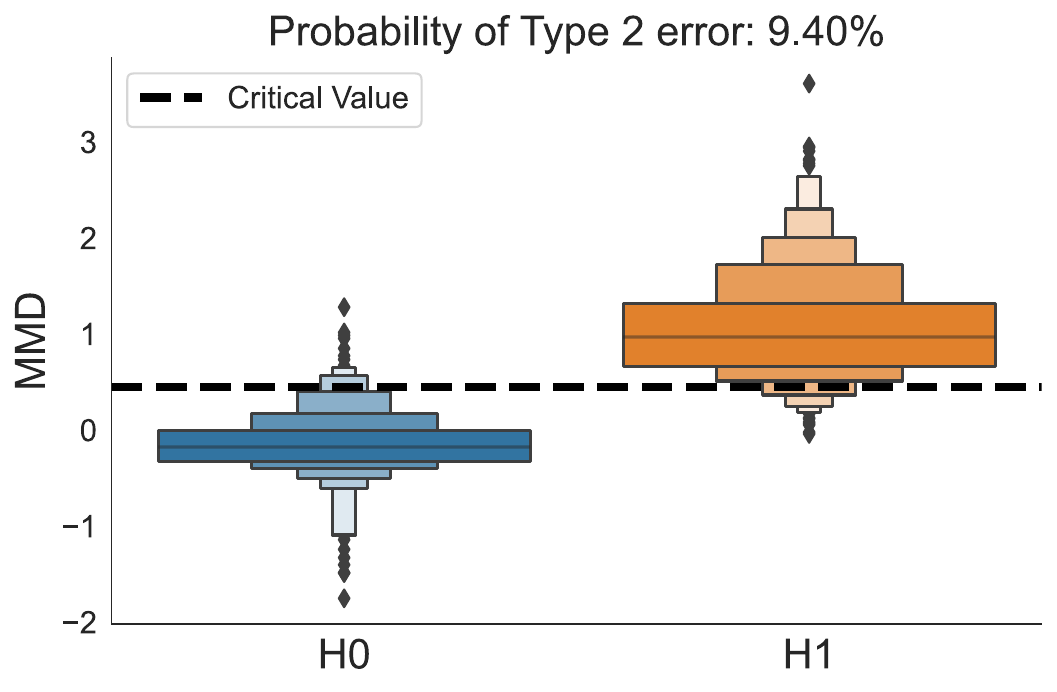}
         \caption{$\sigma_{\text{RBF}} = \sqrt{0.5}$, scaling$=5$.}
         \label{fig:real_world_rbf_scaling_5_leadlag}
     \end{subfigure}
 	\caption{Null and alternative distributions of the $\phi$-MMD after applying lead-lag transformation. $500$ independent simulations were run. Batch size of $128$ and unbiased estimator were used.}
        \label{fig:real_world_leadlag}
\end{figure}

Repeating the same tests after transforming the time series using the lead-lag transform (Fig.\ \ref{fig:real_world_leadlag}), we get a significant reduction in probability of Type 2 errors occurring when scaling was applied after a feature space transformation. Performing a permutation test on the remaining $20\%$ of the data after applying a lead-lag transformation, using an RBF kernel with smoothing parameter of $\sqrt{0.5}$, and applying a scaling of $5$, the hypothesis test concluded that the distributions are not equal. 

\section{Conclusion}

This work focuses on two-sample hypothesis testing between stochastic processes. Whilst the sig-MMD has proven to be an effective tool for two-sample hypothesis testing, careful consideration is needed to fine-tune the test hyperparameters to the samples available. We explored techniques to make the test more robust; namely
\begin{enumerate}
    \item Path scaling;
    \item Lead-lag transformation; 
    \item Standardisation; and
    \item Feature space transformation using the RBF kernel. 
\end{enumerate}
The relationship between the batch size and probability of a Type 2 error occurring was also explored. In addition, we also showed that path scalings do not adversely affect the probability of a Type 1 error occurring. \newline

A contribution made through this work is quantifying the amount of information carried by the level terms of the signature. We showed that the level terms of the $\phi$-MMD incorporate specific distributional properties which, when analysed, can be used to target particular moments using tailored $\phi$-signature kernels. This guides the design of the two-sample test using the $\phi$-MMD. We show that higher-order levels contain information regarding distributional moments which is crucial for reducing Type 2 errors in the two-sample testing framework.\newline 

In summary, the power of the test is a function of the; truncation level, batch size,  feature space (RBF kernel), and $\phi$-signature kernel used. When these are configured appropriately, the two-sample hypothesis test using the $\phi$-MMD as a test statistic is a powerful statistical tool.

\bibliographystyle{abbrv}
\bibliography{ref}

\end{document}